\documentclass[times, twoside, watermark]{zreview-styletex}
\usepackage{blindtext}
\usepackage{cuted}
\usepackage[hang,flushmargin]{footmisc}
\pgfplotsset{compat=1.16}

\newcounter{texexp}
\tcbset{
    texexp/.style={colframe=myblue!80!black, colback=myblue!5!white,
    fonttitle=\small\sffamily\bfseries},
    example/.code 2 args={\refstepcounter{texexp}\label{#2}%
    \pgfkeysalso{texexp,title={Box \thetexexp: #1}}},
}

\usepackage[para]{threeparttable}

\leadauthor{Bonnetto, Qi et al.}
\usepackage{graphicx}
\usepackage{pgfplots}
\usepackage{siunitx} 
\usepackage{epstopdf}
\usepackage{balance}
\usepackage{dsfont}

\usepackage{booktabs}
\usepackage{enumitem}
\usepackage{verbatim}
\usepackage{tabularx}
\usepackage{lipsum,lmodern}
\usepackage{array}
\usepackage[utf8]{inputenc} 
\usepackage[T1]{fontenc}    
\usepackage{hyperref}       
\usepackage{xcolor}         
\usepackage{url}            
\usepackage{booktabs}       
\usepackage{amsfonts}       
\usepackage{nicefrac}       
\usepackage{microtype}      
\usepackage{graphicx}
\usepackage{multirow}
\usepackage{mathrsfs}
\usepackage{pifont}
\usepackage{wrapfig}
\usepackage{cellspace}
\usepackage{nicematrix}
\usepackage{svg}
\usepackage{wrapfig}
\usepackage{fontawesome}

\renewcommand{\cite}[1]{[\citenum{#1}]} 

\newcommand{\twolines}[2]{
\begin{tabular}{@{}c@{}}#1 \\ #2\end{tabular}
}
\newcommand{\twolinesleft}[2]{
\begin{tabular}{@{}l@{}}#1 \\ #2\end{tabular}
}
\definecolor{red_cross}{RGB}{219,20,20}
\definecolor{green_check}{RGB}{20,170,20}
\definecolor{yelloworange}{RGB}{255,174,0}
\definecolor{greentree}{RGB}{26,155,50}
\definecolor{cyanblue}{RGB}{69,191,219}

\newcommand{\cmark}{\textcolor{green_check}{\text{\ding{51}}}}
\newcommand{\xmark}{\textcolor{red_cross}{\text{\ding{55}}}}

\definecolor{lightlightgray}{gray}{0.9}

\newcommand{\threelines}[3]{\begin{tabular}{@{}l@{}l@{}l}#1 \\ #2 \\ #3\end{tabular}}

\newcommand{\fivelines}[5]{\begin{tabular}{@{}l@{}l@{}l@{}l@{}l}#1 \\ #2 \\ #3 \\ #4 \\ #5\end{tabular}}

\begin{document}

\title{EPFL-Smart-Kitchen-30: Densely annotated cooking dataset with 3D kinematics to challenge video and language models}

\shorttitle{EPFL-Smart-Kitchen-30}

\author{{\normalsize Andy Bonnetto\textsuperscript{1,*} ~~~~~~~ Haozhe Qi\textsuperscript{1,*} ~~~~~~~ Franklin Leong\textsuperscript{1} ~~~~~~~ Matea Tashkovska\textsuperscript{1} ~~~~~~~ Mahdi Rad\textsuperscript{2} ~~~~~~~ ~~~~~~~ Solaiman Shokur\textsuperscript{1,3} ~~~~~~~ Friedhelm Hummel\textsuperscript{1,4,5} ~~~~~~~ Silvestro Micera\textsuperscript{1,3} ~~~~~~~ Marc Pollefeys\textsuperscript{2,6} ~~~~~~~ Alexander Mathis\textsuperscript{1,\Envelope}} \\
        \vspace{0.2em}
       {\normalsize 1: École Polytechnique Fédérale de Lausanne (EPFL), Lausanne}
       {\normalsize 2: Microsoft}
       {\normalsize 3: Scuola Superiore Sant'Anna, Pisa}\\
       {\normalsize 4: Swiss Federal Institute of Technology Valais (EPFL Valais), Clinique Romande de Réadaptation, Sion}\\
       {\normalsize 5: University of Geneva Medical School, Geneva}
       {\normalsize 6: Eidgenössische Technische Hochschule (ETH), Zürich} \\
       {\normalsize *: equal contribution} 
       {\normalsize \Envelope: alexander.mathis@epfl.ch}
      }



\maketitle

\begin{abstract}
Understanding behavior requires datasets that capture humans while carrying out complex tasks. The kitchen is an excellent environment for assessing human motor and cognitive function, as many complex actions are naturally exhibited in kitchens from chopping to cleaning. Here, we introduce the EPFL-Smart-Kitchen-30 dataset, collected in a noninvasive motion capture platform inside a kitchen environment. Nine static RGB-D cameras, inertial measurement units (IMUs) and one head-mounted HoloLens~2 headset were used to capture 3D hand, body, and eye movements. The EPFL-Smart-Kitchen-30 dataset is a multi-view action dataset with synchronized exocentric, egocentric, depth, IMUs, eye gaze, body and hand kinematics spanning 29.7 hours of 16 subjects cooking four different recipes. Action sequences were densely annotated with 33.78 action segments per minute. Leveraging this multi-modal dataset, we propose four benchmarks to advance behavior understanding and modeling through 1) a vision-language benchmark, 2) a semantic text-to-motion generation benchmark, 3) a multi-modal action recognition benchmark, 4) a pose-based action segmentation benchmark. We expect the EPFL-Smart-Kitchen-30 dataset to pave the way for better methods as well as insights to understand the nature of ecologically-valid human behavior. Code and data are available at \href{https://github.com/amathislab/EPFL-Smart-Kitchen}{https://github.com/amathislab/EPFL-Smart-Kitchen}. 
\end{abstract}

\section{Introduction}
\label{sec:introduction}

\begin{figure*}[t]
    \centering
    \includegraphics[width = \linewidth]{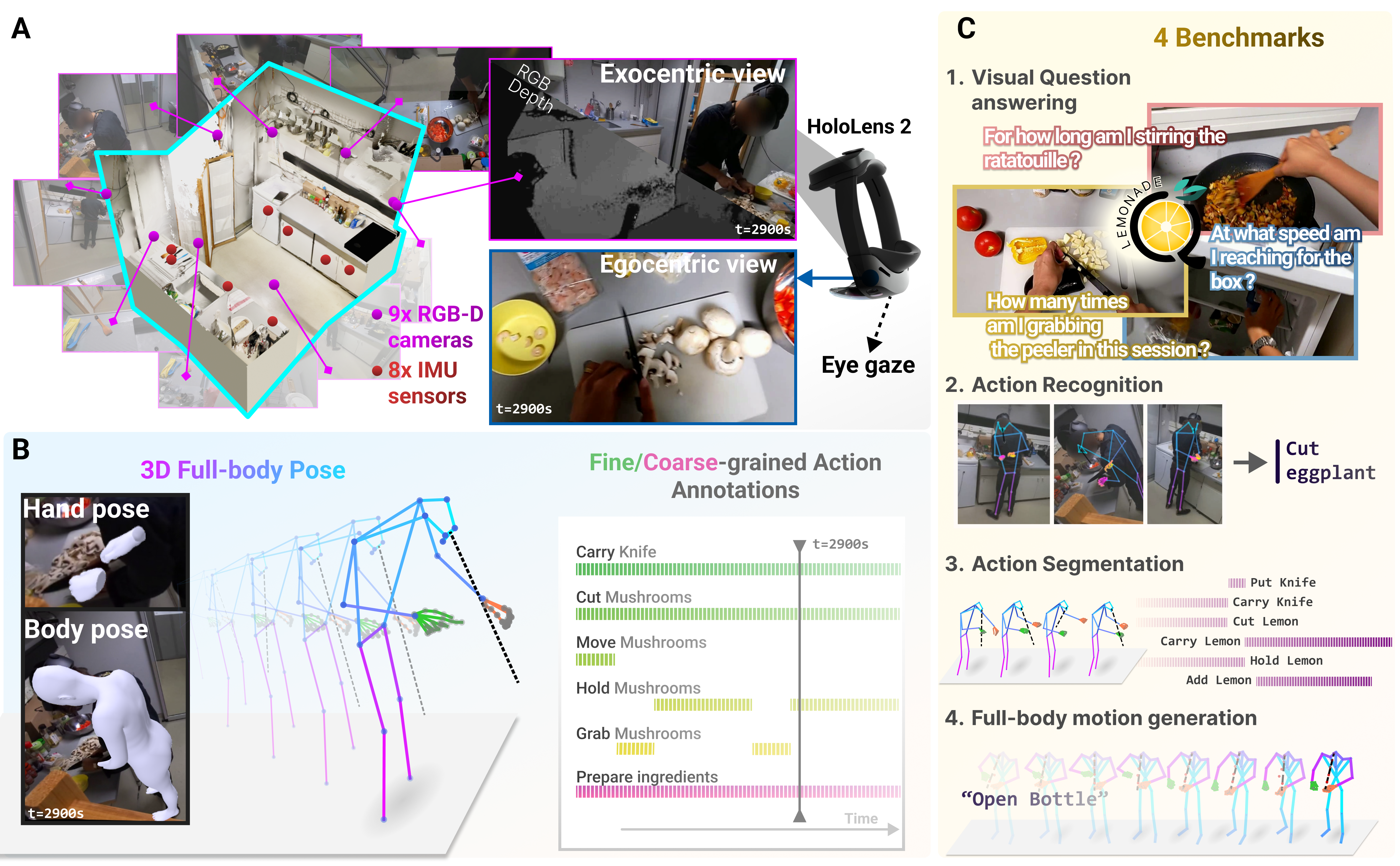}
    \caption{\textbf{The EPFL-Smart Kitchen-30, dataset and benchmarks.} (A) \textbf{Collected data}. 3D kitchen reconstruction, purple points are fixed RGB-D cameras. Subjects cook with a HoloLens~2 headset recording both egocentric videos and eye gaze. (B) \textbf{Extracted data}. 3D body and hand poses are extracted from multiple data sources. Fine-grained and coarse-grained action segments are densely annotated. (C) \textbf{Benchmarks.} We propose four benchmarks based on the EPFL-Smart-Kitchen-30 dataset. A Visual question answering benchmark, an action segmentation benchmark, an action recognition benchmark and a full-body motion generation benchmark.}
    \label{fig:esk_overview}
\end{figure*}

Understanding human behavior is fundamental across multiple domains - from augmented reality\cite{chen2019overview} and robotics~\cite{duan2022survey} to neuroscience~\cite{maselli2023beyond,mathis2024decoding} and neuroengineering~\cite{micera2020advanced}. While we have made significant progress in behavioral analysis through action recognition~\cite{wang2023videomae, li2022uniformer, tong2022videomae, duan2022revisiting}, action segmentation~\cite{van2023aspnet, liu2023diffusion, zhang2022semantic2graph,stoffl2025elucidating} and motion generation~\cite{tevet2022motionclip,zhang2023generating,guo2024momask}, critical gaps remain. Current datasets face a fragmentation problem (Table~\ref{fig:esk_overview}). Existing datasets excel in isolated aspects of behavioral capture, but lack integration. Some datasets advance full-body 3D pose estimation but provide insufficient hand tracking for complex movements. Others offer detailed finger articulations but are limited to constrained environments, missing the crucial full-body context including the global position. Moreover, some datasets fail to capture two essential components of natural behavior: goal-directed actions and eye movements. Human actions are inherently purposeful and guided by visual attention~\cite{hayhoe2005eye}, yet most current datasets do not incorporate these elements, resulting in an incomplete representation of behavior. Comprehensive datasets of full-body, including hand and eye tracking alongside synchronized multi-view video and detailed, hierarchical action annotations are currently missing, and significantly hinder our ability to analyze natural human behavior~\cite{hayhoe2005eye, maselli2023beyond,stoffl2025elucidating}.

We present the EPFL-Smart-Kitchen-30, a dataset that captures humans in authentic cooking scenarios with unprecedented multimodality. It features both egocentric and exocentric perspectives through ten synchronized camera views, providing excellent visual coverage of natural cooking behaviors. EPFL-Smart-Kitchen-30 includes multiple modalities: RGB and depth images, IMU data, eye gazes, and 3D hand/body poses (Figure~\ref{fig:esk_overview}A-B). The EPFL-Smart-Kitchen-30 compares favorably to other datasets (Table~\ref{tab:dataset_comparison}) and promises to advance multimodal fine-grained action understanding. The scale is substantial: 29.7 hours of multi-view, multimodal recordings from 16 participants across 49 complete cooking sessions, from recipe reading to cleanup. The dataset defines 763 fine-grained actions, ensuring dense, hierarchical action annotation and exclusive action definitions. Sessions are densely annotated, yielding 55,361 fine-grained action segments and 4,828 coarse-grained activity segments—about 33 actions per minute.

With the annotated data, we build four benchmarks for action understanding and modeling (Figure~\ref{fig:esk_overview}C).
First, we introduce Lemonade, a novel approach that transforms our ground truth annotations and pose estimations into challenging close-ended question-answer pairs (QA). This benchmark specifically tests the behavioral understanding capabilities of video-language models (VLMs). Second and third, we propose action recognition and segmentation benchmarks that span multiple modalities, providing empirical insights into procedural human behavior. Fourth, we present a full-body motion generation benchmark that highlights EPFL-Smart-Kitchen-30's unique value for generative tasks, demonstrating how our integrated data approach enables more natural and contextually appropriate motion synthesis. In summary, we make the following contributions (Figure~\ref{fig:esk_overview}):

\begin{itemize}
    \item We capture 30 hours of goal-directed cooking behavior from ego-exo perspectives
    \item We densely annotate fine-grained actions and coarse-grained activities.
    \item We propose multimodal behavior understanding (action recognition, segmentation and vision-language question answering) and modeling benchmarks
\end{itemize}

These contributions collectively address the fragmentation problem in behavioral analysis and provide the research community with new possibilities for integrated, context-rich human behavior understanding.

\begin{table*}[t]
    \renewcommand{\arraystretch}{1.}
    \begin{centering}
    \small
    \setlength{\tabcolsep}{2.7pt}

    \begin{NiceTabular}{lcccccccccccccc}[code-before = \rowcolor{lightgray}{16}]
        \toprule
         &\twolines{}{Datasets} & \twolines{Total}{hours} & \twolines{Duration}{(min)} & \twolines{\#}{segments} & \twolines{Seg.}{per min} & \twolines{\# action}{classes} & \twolines{\# Ego/}{Exo}  & \twolines{Body}{PM} & \twolines{Hand}{PM} & \twolines{Eye}{Gaze} & \twolines{}{SA} & \twolines{}{ML} & \twolines{}{DR} & \twolines{}{AP} \\
         \midrule
        \multirow{6}{*}{\rotatebox{90}{\twolines{Video-} {focused}}}
        &Meccano~\cite{ragusa2021meccano}        & 6.9   & 20.7          & 8,857     & 21.4    & 61           & \textcolor{red_cross}{0/1}     & \xmark & \xmark & \cmark & \cmark & \cmark & \cmark & \xmark \\
        &IKEAASM~\cite{ben2021ikea}              & 11.7  & 1.9           & 17,577    & 8.4     & 33           & \textcolor{green_check}{0/3}   & \xmark & \xmark & \xmark & \cmark & \cmark & \cmark & \cmark \\
        &EPIC-100~\cite{damen2022epickitchen}    & 100.0 & 8.6           & 89,977    & 15.0    & 4,053        & \textcolor{red_cross}{1/0}     & \xmark & \xmark & \xmark & \cmark & \cmark & \xmark & \xmark \\
        &EgoExo4D~\cite{grauman2023egoexo4d}     & 180  & 15.3          & 20,406    & 4.5     & 689          & \textcolor{green_check}{1/4-5} & \xmark & \xmark & \cmark & \cmark & \cmark & \xmark & \xmark \\
        &HoloAssist~\cite{wang2023holoassist}    & 166.0 & 4.5           & 184,838   & 18.6    & 1,887        & \textcolor{red_cross}{1/0}     & \xmark & \xmark & \cmark & \cmark & \cmark & \cmark & \cmark \\
        &EgoExo-Fitness~\cite{li2024egoexofit}   & 32    & 1.5           & 6,131     & 4       & 12           & \textcolor{green_check}{0/4}   & \xmark & \xmark & \xmark & \cmark & \cmark & \xmark & \xmark \\
        \midrule
        \multirow{4}{*}{\rotatebox{90}{\twolines{Motion-} {focused}}}
        &AMASS~\cite{mahmood2019amass}           & 43    & 0.22          & 11,451    & 0.22    & -            & -                            & \cmark & \cmark & \xmark & \xmark & \xmark & \xmark & \xmark \\
        &BABEL~\cite{punnakkal2021babel}         & 43    & 0.39          & 28,000    & 10.7    & 250          & -                            & \cmark & \xmark & \xmark & \cmark & \xmark & \xmark & \xmark \\
        &HumanML3D~\cite{guo2022generating}      & 28.6  & 0.12          & 14,616    & 8.5     & -            & -                            & \cmark & \xmark & \xmark & \xmark & \xmark & \xmark & \xmark \\
        &HumanAct12~\cite{guo2020action2motion}  & -     & -             & 1,191     & -       & 34           & -                            & \cmark & \xmark & \xmark & \cmark & \cmark & \cmark & \xmark \\
        \midrule 
        \multirow{5}{*}{\rotatebox{90}{\twolines{Video-motion} {focused}}}
        &Assembly101~\cite{sener2022assembly101} & 41.8  & 7.1           & 84,460    & 33.1    & 1,380       & \textcolor{green_check}{4/8}  & \xmark & \xmark & \xmark & \cmark & \cmark & \xmark & \cmark \\
        &H2O~\cite{kwon2021h2o}                  & 5.5   & 0.33          & 1,000     & 3.0     & 36          & \textcolor{green_check}{1/4}   & \xmark & \cmark & \xmark & \cmark & \cmark & \cmark & \xmark \\
        &MotionX~\cite{lin2023motion}            & 144   & 0.11          & 81,100    & 9.4     & -           & \textcolor{red_cross}{0/1}     & \cmark & \cmark & \xmark & \xmark & \cmark & \xmark & \xmark \\
        &Nymeria~\cite{ma2025nymeria}            & 300   & 15            & -         & -       & -           & \textcolor{green_check}{1/1}   & \cmark & \xmark & \cmark & \xmark & \xmark & \xmark & \cmark \\
        &\textbf{EPFL-Smart-Kitchen-30}             & 29.7  & 35.9          & 60,189    & 33.78   & 768         & \textcolor{green_check}{1/9}  & \cmark & \cmark & \cmark & \cmark & \cmark & \cmark & \cmark \\
        \bottomrule
    \end{NiceTabular}

    \caption{\textbf{Action dataset comparison.} $\#$ indicates "number of" for simplicity. The remaining columns mark following features: parametric model for motion representation (PM), structured actions (SA), markerless video recording (ML), depth recording (DR), and absolute positioning (AP). Note: EgoExo4D~\cite{grauman2023egoexo4d} reports 1422h by summing per camera recording time, where the total activity is 180h, yet only 88.8h annotated with MSCOCO keypoints (numbers from~\cite{ma2025nymeria}).}
    \label{tab:dataset_comparison}
    \end{centering}
\end{table*}

\section{Related work} \label{sec:related_work}

\subsection{Datasets of human behavior}

Many datasets have been proposed that record participants executing purposeful motions~\cite{shahroudy2016ntu, liu2020ntu, punnakkal2021babel}, which were further extended to fitness activities by datasets like EgoExo-Fitness~\cite{li2024egoexofit} and FLAG3D~\cite{tang2023flag3d} to include more complex human body motions. Traditional motion capture approaches focused on isolated movements, offering high controllability but sacrificing critical contextual information. Recent research has shifted toward recording behavior in natural settings, enabling the study of authentic transitions and sequence patterns that characterize genuine human activity. In particular, absolute positioning determines the agent's spatial location and facilitates the analysis of its interactions with the environment~\cite{ma2025nymeria}. Assembly-based datasets collect structured object interactions~\cite{ben2021ikea, wang2023holoassist, alayrac2016unsupervised, sener2022assembly101}, but by using a greater variety of actions and natural environments, cooking is getting popular for building action datasets such as EPIC-KITCHENS-100~\cite{damen2022epickitchen}, Humans in kitchens~\cite{tanke2024humansinkitchens} or certain sequences of the large EgoExo4D~\cite{grauman2023egoexo4d} dataset. The EPFL-Smart-kitchen-30's dense annotations suit the characterization of body and hand movement transitions and distinguish themselves with their unambiguous and rich action descriptions (Table~\ref{tab:dataset_comparison}).

Language can flexibly describe behavior, and VLMs promise to capture that richness. The general video understanding of VLMs has been evaluated with exhaustive benchmarks such as MVBench~\cite{li2024mvbench} and Video-MME~\cite{fu2024videomme}. More specific challenges subsequently developed to tackle long-term understanding~\cite{zhou2024mlvu, mangalam2023egoschema} and egocentric video understanding~\cite{mangalam2023egoschema, koyejo2022egotaskqa}. In the case of behavior understanding, ActivityNet-QA~\cite{yu2019activitynet} and NExT-QA~\cite{xiao2021nextqa} evaluate the causal and temporal abilities of VLMs. While subsets of certain benchmarks~\cite{cores2024tvbench, li2024mvbench, zhou2024mlvu} contain questions related to motion, they mostly focus on understanding behavior at the event level. EPFL-Smart-Kitchen-30 enables a fundamentally different approach. Our Lemonade benchmark introduces questions that specifically probe the understanding of human kinematics and fine-grained behavioral details that previous datasets simply cannot address. By leveraging our multimodal data integration, we can evaluate models on their ability to reason about body movements and hand-object interactions (Figure~\ref{fig:lemonade}).

 \subsection{Models for behavior understanding}
 
The ability to predict movement patterns provides a valuable approach to understanding behavior. Movement can be captured in various forms, including video recordings, pose estimation data, and IMU recordings. Improvements in deep learning models together with their increase in computational power levels have led to the development of many multi-view, multimodal action understanding algorithms \cite{shah2023multi, wang2019generative, zhao2023learning, chalk2024tim, xiao2020audiovisual, shamil2024utility}. \citeauthor{shah2023multi}~\cite{shah2023multi} leverage contrastive learning to align the feature spaces from different views. \citeauthor{wang2019generative}~\cite{wang2019generative} use an adversarial generative network to constrain RGB and depth modality information. HandFormer~\cite{shamil2024utility} combines 3D hand poses and RGB frames together for action recognition. LaViLa~\cite{zhao2023learning} learns video representations from pre-trained large language models. TIM~\cite{chalk2024tim} designs time interval encodings to incorporate visual and audio events. Despite progress, current methods are limited in views and modalities, partially due to the lack of large-scale multi-view, multimodal action datasets. With our EPFL-Smart-Kitchen-30 dataset, we set up multi-view, multimodal action understanding benchmarks taking and comparing exocentric videos, egocentric videos, full-body pose estimations, and eye gaze modalities as input, with the possibility to also include depth videos and IMU recordings. 

Another approach for behavior understanding is through the ability to generate movement of a target behavior. Recently, text-to-motion generation gained a lot of attention~\cite{tevet2022motionclip,chen2023executing, zhang2023generating, guo2024momask, tevet2023human, pinyoanuntapong2024mmm, zhang2023remodiffuse}. We propose a novel semantic text-to-motion generation benchmark that considers full-body pose representations, including eye gaze, for situated motion generation. This contrasts with the commonly used KIT~\cite{plappert2016kit} and HumanML3D~\cite{guo2022generating}, which do not incorporate hand models or gaze information.

By integrating language, VLMs provide more flexible ways to understand behavior. VideoLLaMA3~\cite{zhang2025videollama} captures fine-grained details and temporal dynamics in videos through its dynamic resolution mechanisms and advanced positional embedding strategies, whereas Qwen2.5-VL~\cite{bai2025qwen25} and Intern VL2.5~\cite{chen2024expanding} better integrate multimodal inputs. Specific tasks such as long-term video understanding usually rely on video compression~\cite{li2024llava, shen2024longvu,li2024llamavid, Lin2023vila} or on extending their context length~\cite{zhang2024longcontext, weng2024longvlm, chen2024longvila, liu2025worldmodel}. We challenge these models to operate beyond their conventional performance by proposing a benchmark that leverages behavioral context and kinematics.

\section{The EPFL-Smart-Kitchen-30}

\begin{figure*}[t]
    \centering
    \includegraphics[width = \linewidth]{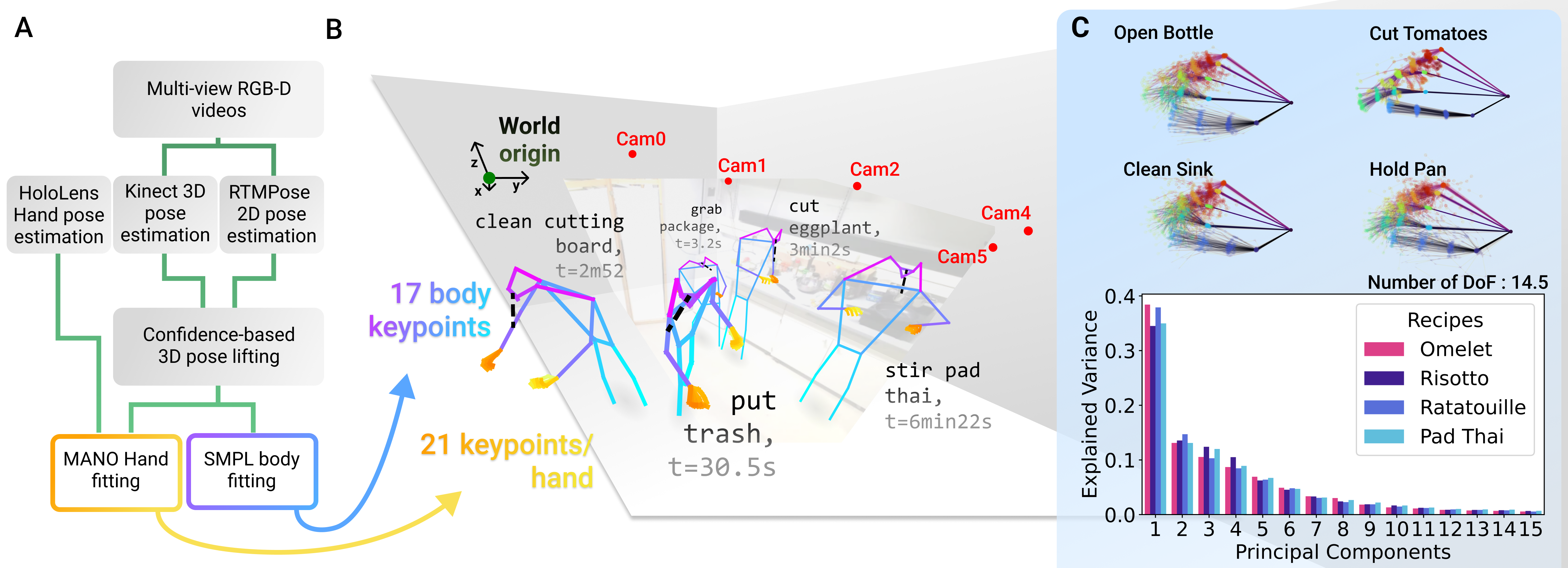}
    \caption{\textbf{Full-body 3D pose estimation} (A) Pipeline for 3D pose estimation (B) Poses and camera positions are defined relative to a global coordinate system comparable in the same environment. We illustrate several reprojected 3D poses on camera 6. (C) Characterization of right-hand poses to show the captured kinematic diversity: (Top) examples of hand poses for four actions, (Bottom) Number of principal components necessary to capture the right-hand poses during cooking for the Exo-Hand in each recipe, around 14 degrees of freedom (DoF) are necessary to explain 95\% of the variance.}
    \label{fig:pose_example_richness}
\end{figure*}

Here we introduce the EPFL-Smart-Kitchen-30 dataset, which features multi-view, multimodal data of human cooking with fine-grained and coarse-grained action annotations (Figure~\ref{fig:esk_overview}B). We will describe the setup (Sec.~\ref{subsec:content}) and the data collection procedure (Sec.~\ref{subsec:datacollection}). Then, we illustrate the 3D pose regression (Sec.~\ref{subsec:3dpose}) and detail the action annotation characteristics (Sec.~\ref{subsec:annotations}).

\subsection{EPFL-Smart-Kitchen setup} \label{subsec:content}

Capturing multi-view, multimodal data is a challenging task, requiring the synchronization and calibration of multiple sensors. To capture naturalistic cooking behaviors, we built the EPFL-Smart-Kitchen, a fully functional kitchen with appliances and utensils. Cooking materials including pots, pans, and other utensils were provided to the subjects along with the ingredients and spices necessary for preparing the recipe (see Supp. Mat.). 

To minimally affect the subjects' natural movements while capturing multimodal information, we installed nine Microsoft Kinect Azure RGB-D cameras~\cite{kinectazure} at strategic points inside the kitchen, four focusing on the global exocentric view and five focusing on local exocentric views (counters, stove, and sink, see Figure~\ref{fig:esk_overview}A).
We additionally equipped the kitchen with eight IMU sensors on the frequently used equipment (e.g., fridge door, five cupboard doors, knife, and spatula). Subjects wore a Hololens~2 headset~\cite{ungureanu2020hololens}, a mixed-reality headset that can capture egocentric views and eye gaze data under global calibration. We synchronized all devices using audio signals and a trigger, and calibrated all the cameras (see Supp. Mat.). 

\subsection{Data collection procedure}  \label{subsec:datacollection}

To capture realistic cooking scenarios, subjects prepared a meal from reading a recipe to cleaning up, leading to significantly longer recordings than most existing action datasets (Table \ref{tab:dataset_comparison}). We recruited 16 subjects (four males and twelve females, two left-handed, ages 20-46) to cook for up to five sessions in the EPFL-Smart-Kitchen (see Supp. Mat.).  During each session, subjects are asked to follow one of four different recipes (omelet, pad thai, risotto, ratatouille), adapted to their preferences and requirements. Overall, we recorded and processed 29.7h of cooking experiments, corresponding to 3,207,600 frames per camera for 49 cooking sessions. All procedures were approved by the EPFL-Ethical Board. Subjects' faces are anonymized across all videos to address privacy concerns. All subjects consented and were informed about the ethics (see Supp. Mat.).

\subsection{Estimation of 3D motions} \label{subsec:3dpose}

We placed the cameras so that both the body and the hands of the participants are visible from at least three angles. Four cameras captured global body information, while five cameras captured local hand information. Using multi-view RGB-D video, we conduct body/hand mesh fitting and tracking using all 10 camera views, extracting 2D and 3D pose information from each view with existing pose estimation tools. Specifically, we extract 2D body and hand poses using RTMPose~\cite{jiang2023rtmpose}, available in DeepLabCut v3~\cite{mathis2018deeplabcut}, 3D body poses and tracklets using Kinect body tracking SDK~\cite{kinectbody}, and 3D hand poses using HoloLens~2 hand tracking toolkit~\cite{holohand}. We lift 2D pose to 3D poses and fit the SMPL~\cite{SMPL-X:2019} body mesh by minimizing the 3D joint, 2D reprojection, temporal smoothing, and regularization loss as well as the hand 3D joint loss to fit the MANO hand mesh~\cite{MANO:SIGGRAPHASIA:2017} (Figure \ref{fig:pose_example_richness}A and Suppl. Mat.). The average absolute error compared to triangulated manually-annotated 2D poses is $6.22 cm \pm 5.16cm$ and $3.30cm \pm 5.12cm$ for the body and hand respectively, our margins are comparable with those of ~\cite{grauman2023egoexo4d, kwon2021h2o} (Supp. Mat.).

To illustrate the richness of the captured movement data (Figure~\ref{fig:pose_example_richness}B-C), we estimate the number of degrees of freedom (kinematic synergies) in pose space based on a common method in neuroscience~\cite{todorov2004analysis,santello1998postural,chiappa2024acquiring}. We found that the dataset exhibits a large number of degrees of freedom (Figure~\ref{fig:pose_example_richness}C), which foreshadows the potential for human behavior studies.

\subsection{Annotation of fine-grained actions and coarse activities} \label{subsec:annotations}

\begin{figure*}[t]
    \centering
    \includegraphics[width = \linewidth]{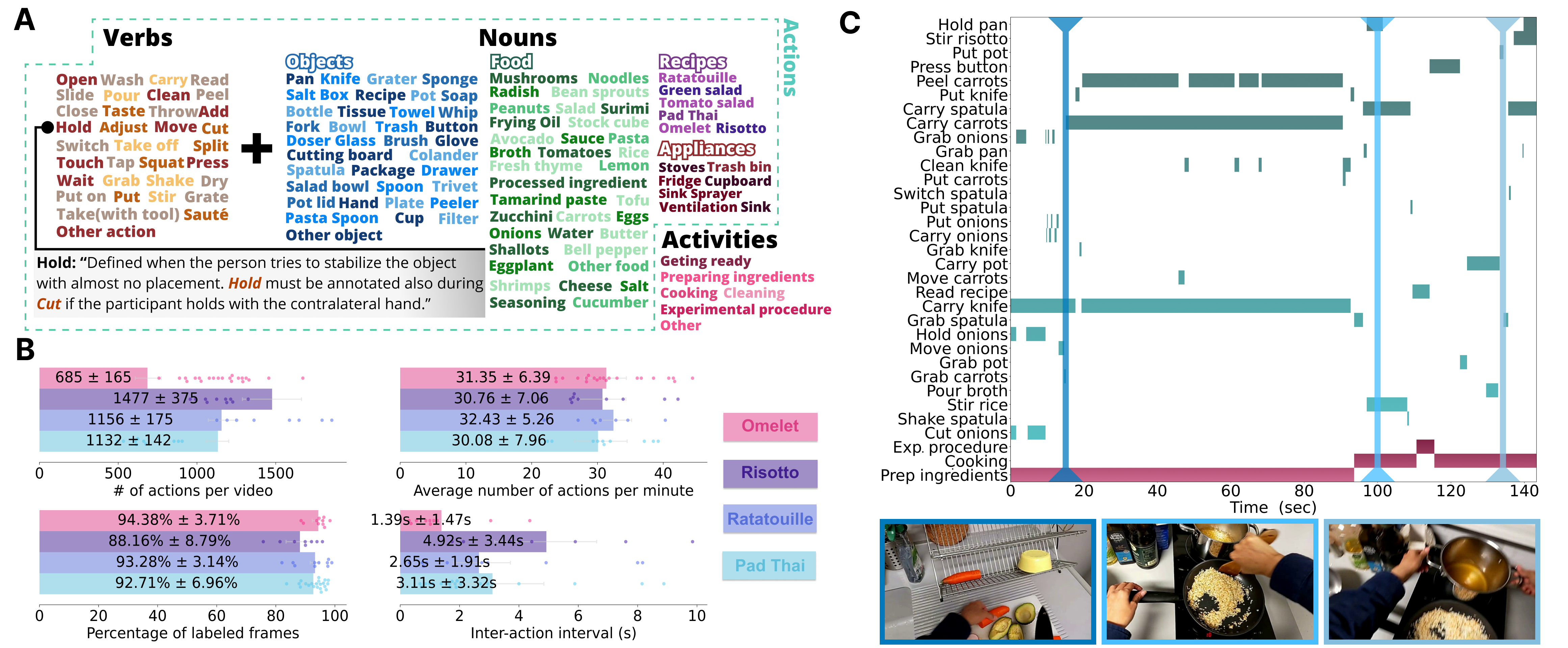}
    \caption{\textbf{Hierarchical action annotations}: (A) List of verbs, nouns and activities used for action annotation, each action (verbs) is specifically defined as shown for "Hold". (B) Statistics of annotation segments for fine-grained actions. (C) Example ethogram with selected egocentric frame to illustrate the richness of the actions (turquoise) and activity (pink) annotations comprising short and long segments that can overlap.}
    \label{fig:action_annotations}
\end{figure*}

The annotated action classes were defined with the following considerations. Firstly, many contemporary datasets (e.g.,\cite{damen2022epickitchen, wang2023holoassist}) tend to let the annotators freely describe the actions and then post-hoc group the actions based on action similarity. This might lead to different names for similar actions (e.g., \textit{pour} and \textit{fill}~\cite{damen2022epickitchen}) and thus introduce ambiguity. We instead curated a set of verbs and nouns. We annotated with temporal overlaps between actions. For example, when labeling \textit{cut tomato}, we also label \textit{carry knife} and may label \textit{hold tomato}. This enriches the annotation at a given timestep and attempts to reduce ambiguity. Based on the above rationale, we define 33 verbs and 79 nouns, which compose 763 fine-grained actions. Each verb is defined by a rule-based description intended to prevent confusion (see example in Figure~\ref{fig:action_annotations}A and Supp. Mat.). During the annotation procedure, we asked annotators to watch videos and annotate the start and stop times of actions. To define the behavioral contexts for each action, six coarse-grained exclusive activities were annotated, summing up to 4,828 segments. Thus, behavior is annotated in a hierarchical fashion~\cite{anderson2014toward,stoffl2025elucidating,gabeff2025mammalps}.

The quality and reliability of annotations were validated following the protocol outlined in the Supp. Mat. In total, 60,189 action segments were annotated, resulting in 33.78 action segments per minute (Figure \ref{fig:action_annotations}B). The richness of the action annotation is demonstrated by a large variety of action lengths (from 1 second to 100 seconds, Figure \ref{fig:action_annotations}C). Overall, the different views and modalities contribute to fine-grained action understanding in different aspects: 1) RGB frames focus on coarse-grained information while depth frames rather focus on geometric aspects; 2) the egocentric view and the eye gaze data captured from the HoloLens~2 are related to (part of) what the subject sees; 3) global exocentric views and the body poses capture the overall context; 4) local exocentric views, hand poses, and tool IMU data capture fine-grained movements and hand-object interactions.

\section{Multimodal action and motion understanding benchmarks} \label{sec:benchmarks}

Cooking involves many different actions that are sequenced in a goal-directed fashion to achieve a tasty outcome. In each experimental session, subjects go from reading the recipe and preparing the ingredients to creating the dish and ultimately cleaning up. To make progress towards analyzing such complex human behavior, we created four behavior understanding benchmarks (Figure~\ref{fig:esk_overview}C). Two benchmarks focus on multimodal behavior analysis: action recognition and action segmentation. These benchmarks are complemented by a full-body behavior synthesis benchmark (motion generation). Furthermore, we designed a question-answering benchmark (Lemonade) to understand human cooking behavior. Lemonade is structured for zero-shot evaluation. For the other benchmarks, we split the sessions into train, validation, and test sets. The training/validation sets are split into 26/7 sessions chosen so that every recipe is present in the validation set and to balance the number of rare action segments in both sets. The test set is composed of 16 sessions which also include new subjects (see Supp. Mat.). The curated dataset used for the benchmark excludes actions with less than 3 instances and is composed of 31 verbs, 78 nouns, and 581 actions together with six activities. 

\begin{figure*}[t]
    \centering
    \includegraphics[width=\linewidth]{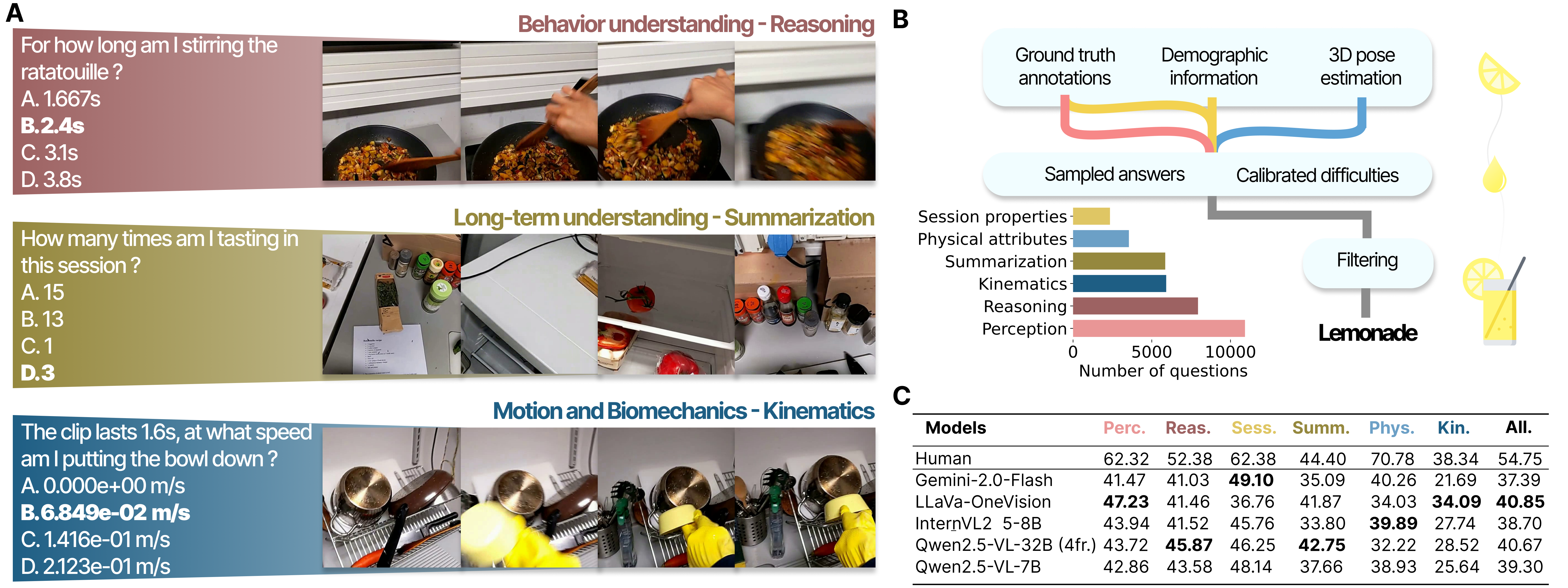}
    \caption{\textbf{Lemonade}: (A) Examples of video question pairs for each category. More examples in Supp. Mat. (B) Questions are designed from ground truth annotations. We also show the distribution of questions per subcategory. (C) Accuracy of VLMs (8 frames) on the lemonade benchmark per category; chance level is 25\%. Detailed results in Supp. Mat.}
    \label{fig:lemonade}
\end{figure*}

\subsection{Lemonade: Language models Evaluation of MOtion aNd Action-Driven Enquiries}

\noindent{\bf Rationale.} Video-Language Models (VLMs) exhibit remarkable potential for understanding human behavior~\cite{wang2024qwen2,zhang2024lmmseval,li2024llava,ye2025llavaction}. They raise intriguing questions: Can they accurately predict preceding or subsequent actions in behavioral sequences? Are they able to infer long-term behavioral patterns from just a few frames? Furthermore, can their general knowledge enable precise distance and velocity estimations from video data alone? EPFL-Smart-Kitchen-30 provides the ideal testbed to explore these fundamental questions about machine understanding of natural human behavior (Figure~\ref{fig:lemonade}A).
Thus, we introduce Lemonade: \textbf{L}anguage models \textbf{E}valuation of \textbf{MO}tion a\textbf{N}d \textbf{A}ction-\textbf{D}riven \textbf{E}nquiries. Lemonade consists of 36,521 closed-ended QA pairs linked to egocentric video clips, categorized in three groups and six subcategories (Figure~\ref{fig:lemonade}B). 18,857 QAs focus on behavior understanding, leveraging the rich ground truth behavior annotations of the EPFL-Smart Kitchen to interrogate models about perceived actions (Perception) and reason over unseen behaviors (Reasoning). 8,210 QAs involve longer video clips, challenging models in summarization (Summarization) and session-level inference (Session properties). The remaining 9,463 QAs leverage the 3D pose estimation data to infer hand shapes, joint angles (Physical attributes), or trajectory velocities (Kinematics) from visual information. The Lemonade framework (Figure \ref{fig:lemonade}B) is designed to generate millions of unique QA pairs by combining various video clips, question formats and answer types. More examples and details on QA design can be found in the Supp. Mat.

\noindent{\bf Baselines and Metrics.} Based on state-of-the-art results from other recent benchmarks~\cite{tu2025favorbench}, we evaluated three open-source and one closed-source VLM SoTA models, namely InternVL2.5~\cite{chen2024expanding}, LLaVA-OneVision~\cite{li2024llava}, Qwen2.5-VL~\cite{bai2025qwen25} and Gemini 2.0 Flash~\cite{google2024gemini}. To ensure consistent evaluation and enable future comparisons, all models are evaluated using lmms-eval~\cite{zhang2024lmmseval}, where Lemonade was implemented as a new evaluation task. To interpret the results, we manually answered 1,662 questions. As commonly done for question answering, we evaluate the model performance as average accuracy~\cite{fu2024videomme, li2024mvbench, xiao2021nextqa, yu2019activitynet}.

\noindent{\bf Results.} Different VLMs achieved high accuracy in identifying ongoing actions and activities, as well as predicting immediate next and previous actions. However, Lemonade exposed critical limitations in VLMs to predict general context information, distances, timings, and body kinematics (Figure \ref{fig:lemonade}C). Merely relying on visual input and language proved insufficient for these precise kinematic estimations, which require accurate frame timing and depth reference data. To overcome these challenges, future models could benefit from explicitly integrating additional modalities, which is becoming possible with multimodal language models~\cite{lucas2022posegpt,feng2024chatpose}.

\subsection{Action recognition benchmark}

{\bf Rationale}. Given a trimmed action segment, the action recognition model needs to predict the corresponding action class~\cite{wang2023videomae, li2022uniformer, tong2022videomae, duan2022revisiting}. We built a fine-grained action recognition benchmark for 763 classes with a long-tailed distribution and allow different data types as input. Specifically, we formulated flexible masked auto-encoding baselines taking the egocentric view, one exocentric view, the 3D body poses, the 3D hand poses, and the eye gaze rays as input and compared different combinations of those data sources.

\noindent{\bf Baselines and Metrics.} 
We adapted VideoMAE~\cite{tong2022videomae,gabeff2025mammalps} model into a multi-modal MAE, enabling it to take multi-view videos, 3D poses, and eye gazes as inputs (see Supp. Mat.). Like others, we evaluate the model performance by Top-1 and Top-5 accuracy for verb, object, and action classes ~\cite{wang2023holoassist, sener2022assembly101}. Considering that the number of action samples across different actions has a long-tail distribution, we selected the top 180 most frequent actions as head actions and report the performance for both head actions and tail actions separately.

\begin{table*}[h!]
\centering
\small

\begin{NiceTabular}{Wl{50pt}Wc{36pt}Wc{36pt}Wc{36pt}Wc{0pt}Wc{36pt}Wc{36pt}Wc{36pt}Wc{0pt}Wc{36pt}Wc{36pt}Wc{36pt}}
\toprule
\multirow{2}{*}{Modalities}
 & \multicolumn{3}{c}{All Classes Accuracy Top1/5} && \multicolumn{3}{c}{Head Classes Accuracy Top1/5} && \multicolumn{3}{c}{Tail Classes Accuracy Top1/5} \\
\cmidrule{2-4} \cmidrule{6-8} \cmidrule{10-12}
 &  Action &  Verb &  Noun && Action &  Verb &  Noun && Action & Verb & Noun \\
\midrule
\faCameraRetro                                                & 37.51/62.94 & 57.72/92.18 & 52.03/79.05 && 41.12/67.00 & 59.74/93.36 & 55.62/82.11 && 16.64/39.51 & 46.06/85.35 & 31.27/61.38 \\
\faCameraRetro \faChild                                       & 37.87/63.56 & 58.90/93.59 & 52.56/79.64 && 41.55/67.67 & 60.56/94.66 & 56.43/82.84 && 16.66/39.85 & 49.28/87.44 & 30.19/61.11 \\
\faCameraRetro \faPlayCircle                                  & 37.57/63.13 & 58.38/92.92 & 52.29/78.45 && 41.14/67.31 & 60.43/93.90 & 55.90/81.62 && 16.97/39.00 & 46.54/87.24 & 31.46/60.15 \\
\faChild \faHandPaperO                                        & 11.80/25.49 & 38.67/78.80 & 19.83/42.23 && 13.55/28.77 & 39.87/80.65 & 22.31/46.29 && 1.70/6.57 & 31.77/68.15 & 5.49/18.79 \\
\faCameraRetro \faChild \faHandPaperO                         & 38.31/64.75 & 60.41/93.78 & 52.51/79.91 && 41.83/68.89 & 61.96/94.79 & 56.27/83.14 && 17.97/40.90 & 51.45/87.94 & 30.81/61.27 \\
\faCameraRetro \faChild \faHandPaperO \faEye                  & 37.35/62.34 & 60.78/93.04 & 50.58/77.30 && 40.91/66.05 & 62.20/93.92 & 54.15/80.48 && 16.85/40.97 & \textbf{52.55}/87.94 & 30.02/58.99 \\
\faCameraRetro \faPlayCircle \faChild \faHandPaperO \faEye    & 37.49/62.76 & \textbf{61.04}/93.59 & 50.94/77.52 && 41.09/66.50 & 62.53/94.51 & 54.75/80.57 && 16.66/41.21 & 52.42/\textbf{88.29} & 28.95/59.91 \\
\faCameraRetro \faGears(\faHandPaperO$\times$ \faPlayCircle)  & \textbf{40.03/67.01} & 60.80/\textbf{94.65} & \textbf{55.38/82.58} && \textbf{43.60/71.25} & \textbf{62.60/95.76} & \textbf{59.00/85.62} && \textbf{19.44/42.52} & 50.41/88.25 & \textbf{34.48/65.02} \\
\bottomrule
\end{NiceTabular}
\caption{\textbf{Fine-grained action recognition benchmark results from pretrained model.} \faCameraRetro : egocentric view, \faPlayCircle : global exocentric view, \faChild : 3D body pose, \faHandPaperO : 3D hand pose, \faEye : eye gaze, \faGears(\faHandPaperO$\times$multiple \faPlayCircle) : hand cropped videos. Combining modalities has the potential to increase the performance. Our best results are achieved by cleverly merging these modalities together.}
\label{tab:recognition_baseline}
\end{table*}

\begin{table*}[h!]
    \centering
    \small
    \renewcommand{\arraystretch}{0.7}
    \begin{NiceTabular}{Wl{45pt} Wc{12pt}Wc{12pt}Wc{12pt}Wc{12pt}Wc{12pt}Wc{12pt}Wc{0pt}Wc{12pt}Wc{12pt}Wc{12pt}Wc{12pt}Wc{12pt}Wc{12pt}Wc{0pt}Wc{12pt}Wc{12pt}Wc{12pt}Wc{12pt}Wc{12pt}Wc{12pt}}
    \toprule
     & \multicolumn{6}{c}{Verbs} && \multicolumn{6}{c}{Nouns} && \multicolumn{6}{c}{Activity} \\
    \cmidrule{2-7} \cmidrule{9-14} \cmidrule{16-21}
                & \faChild    & \faHandPaperO   & \faEye     & \faChild \faHandPaperO   & \twolines{\faChild \faHandPaperO}{\faEye} &\twolines{\faChild \faHandPaperO }{\faEye \faCameraRetro} && \faChild   & \faHandPaperO     & \faEye   & \faChild \faHandPaperO & \twolines{\faChild \faHandPaperO}{\faEye} & \twolines{\faChild \faHandPaperO }{\faEye \faCameraRetro} && \faChild    & \faHandPaperO       & \faEye     & \faChild \faHandPaperO        & \twolines{\faChild \faHandPaperO}{\faEye} & \twolines{\faChild \faHandPaperO }{\faEye \faCameraRetro} \\
    \midrule
     MS-TCN3            & 18.1 & 20.2 & 11.7 & 20.9 &  21.1 & 30.1 && 10.6   & 13.4  & 7.6  &  15.6  &  11.3  & 31.2 && 51.8 & 58.6 & 31.9 & 54.4 & 58.5 & \textbf{72.9}\\
     C2F-TCN*           & 18.8 & 20.1 & 12.2 & 22.1 &  22.2 & 34.6 && 12.0   & 14.3  & 7.9  &  16.1  &  10.8  & \textbf{35.2} && 54.5 & 55.4 & 41.3 & 61.8 & 61.2 & 72.2\\
     C2F-Transf.        & 19.9 & 22.4 & 13.1 & 22.8 &  22.2 & \textbf{35.0} && 11.1   & 12.9  & 7.8  &  13.4  &   9.2  & 29.0 && 51.2 & 56.9 & 38.8 & 62.1 & 59.9 & 70.5\\
     EDTCN*             & 19.6 & 23.0 & 11.9 & 22.1 &  25.2 & 34.3 && 11.9   & 11.2  & 7.1  &  12.3  &  11.9  & 24.3 && 49.0 & 53.5 & 32.0 & 53.1 & 54.2 & 71.0\\
    \bottomrule
    \end{NiceTabular}
    \caption{\textbf{F1 scores for action segmentation benchmark.} \faChild : 3D body pose, \faHandPaperO : 3D hand pose, \faEye : eye gaze, \faCameraRetro : egocentric view. *models modified to use pose as input data instead of image features.}
    \label{tab:ActSegMain}
\end{table*}

\noindent{\bf Results}. Simple concatenation of all modalities as inputs yielded slightly better performance compared to single video modality (Table \ref{tab:recognition_baseline}). Meanwhile, transfer learning from a ViT model trained on EPIC-KITCHENS-100~\cite{damen2022epickitchen} boosted the performance for baselines with visual inputs and reduced the performance for pose only. Furthermore, adding (hand and body) pose information to the video-based models yielded better overall performance, which mainly benefits from the verb prediction improvements. However, simply concatenating tokens from different modalities barely improved the performance. To efficiently utilize multiple modalities without significant computational cost, we integrate egocentric view, multi-exocentric views, and hand pose data (\faGears) (see Supp. Mat.) to boost the performance by 21.6\% when trained from scratch (see Supp. Mat.) and 6.3\% when trained from the pretrained model, over the egocentric-only model. Overall, we hope this benchmark will inspire the community to create models that can effectively use multiple modalities.

\subsection{Action segmentation benchmark}

{\bf Rationale}. Given an untrimmed video, action segmentation requires the model to predict one or multiple action classes for every frame~\cite{van2023aspnet, liu2023diffusion, zhang2022semantic2graph,stoffl2025elucidating}. Given the absence of popular (and comprehensive) action segmentation benchmarks from 3D pose data (Sec. \ref{sec:related_work}), we built an action segmentation benchmark that compares the impact of different input data (body, hand, eyes, video features). One might expect that actions such as moving through the kitchen will be better predicted from the body, while motions like cutting require hand pose keypoints. Therefore, we used combinations of body pose (\faChild), hand poses (\faHandPaperO) and eye gazes (\faEye) to form the input as they are computationally more efficient than deep visual features. We additionally compared the performance when using video features from VideoMAE as input.

\noindent{\bf Baselines and Metrics.} We considered state-of-the-art models on RGB-based action segmentation tasks (Breakfast~\cite{kuehne2014language} and 50Salads~\cite{stein2013combining}) and adapted these architectures to work directly on pose estimation data (vs deep visual features): MS-TCN++~\cite{li2020ms-tcn++}, EDTCN~\cite{Lea_2017_CVPR} and C2F-TCN~\cite{singhania2022iterative}. Based on these models, we adapted architectures that have been optimized on pose estimation data: MS-TCN3 and C2F-Transformer. We used kinematic features and VideoMAE~\cite{tong2022videomae} features as input to the models (see Supp. Mat.). Each action is evaluated separately using standard metrics in action segmentation (Frame-wise F1, F1@50, edit distance) and ultimately averaged over action groups.

\noindent{\bf Results}. We observed that the benchmark is challenging for current action segmentation algorithms (Table~\ref{tab:ActSegMain} and Supp. Mat.). Exo-Body performs similarly to Exo-Hand with a slight improvement for Exo-Hand albeit for different behaviors (see Supp. Mat.). We note that video information provided a boost in performance. These baselines showed an F1-score of 35.2\% for verbs and 35.0\% for nouns, highlighting significant potential for future advancements.

\subsection{Situated full-body motion generation benchmark}

{\bf Rationale.} With the diverse actions and motions in EPFL-Smart-Kitchen-30, our motion generation benchmark has three key innovations over the commonly used KIT~\cite{plappert2016kit} and HumanML3D~\cite{guo2022generating} benchmarks. Existing motion generation benchmarks mainly focus on broad daily activities, sports, and dance, whereas the EPFL-Smart-Kitchen-30 contains hundreds of fine-grained behaviors. We extend motion generation beyond the body to include hands and eye gaze, defining it as full-body motion generation. Additionally, we provide egocentric visual features alongside action text as the condition, allowing situated motion generation.

\begin{figure*}[h]
    \centering
    \includegraphics[width=\linewidth]{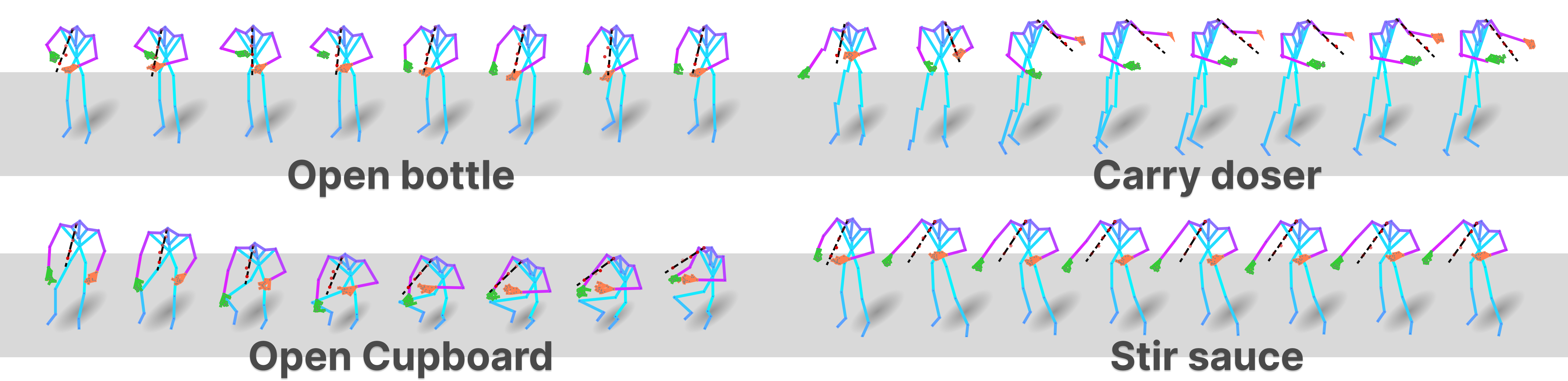}
    \caption{\textbf{Qualitative motion generation samples} from MoMask trained with Verb-Noun action pairs. The model creates realistic hand, body and eye gaze (dashed line).}
    \label{fig:motion_generation_examples}
    \vspace{10pt}
\end{figure*}

\begin{table*}[h]
    \centering
    
    \small
    
    \begin{NiceTabular}{Wl{50pt}Wl{40pt}Wl{45pt}Wc{36pt}Wc{36pt}Wc{36pt}Wc{36pt}Wc{36pt}Wc{36pt}Wc{36pt}}
    \toprule
    Condition& Vocab.          &  Model   &  T1 ↑  & T2 ↑  & T3 ↑  & FID ↓ & DIV ↑ & MM ↑  & MMd ↓\\
     \midrule
     \multirow{6}{*}{Text}
    &\multirow{3}{*}{Verb}     &  T2M-GPT & 0.254  & 0.432 & 0.567 & 2.640 & 7.288 & 1.555 & 3.379  \\
    &                       &  MARDM-SiT  & 0.262  & 0.473 & 0.637 & 2.242 & 7.590 & 1.849 & 4.155  \\
    &                          &  Momask  & 0.306  & 0.508 & 0.652 & 3.124 & \textbf{8.347} & \textbf{2.940} & \textbf{2.048}  \\
    &\multirow{3}{*}{Actions}  &  T2M-GPT & 0.271  & 0.434 & 0.542 & 2.378 & 7.031 & 0.852 & 4.015 \\
    &                       &  MARDM-SiT  & 0.320  & 0.548 & 0.650 & 1.230 & 7.704 & 1.427 & 4.103  \\
    &                          &  Momask  & \textbf{0.372}  & \textbf{0.566} & \textbf{0.683} & 0.930 & 7.859 & 1.641 & 3.407 \\
    \midrule
    \multirow{6}{*}{Text-Image}
    &\multirow{3}{*}{Verb}     &  T2M-GPT & 0.174  & 0.295 & 0.387 & 2.415 & 6.558 & 0.822 & 4.524  \\
    &                       &  MARDM-SiT  & 0.187  & 0.355 & 0.466 & 1.498 & 6.579 & 1.581 & 4.691  \\
    &                          &  Momask  & 0.197  & 0.333 & 0.436 & 0.858 & 6.696 & 1.665 & 4.121  \\
    &\multirow{3}{*}{Actions}  &  T2M-GPT & 0.243  & 0.398 & 0.506 & 1.982 & 6.633 & 0.717 & 4.125 \\
    &                       &  MARDM-SiT  & 0.255  & 0.469 & 0.597 & 0.917 & 7.176 & 1.322 & 4.725  \\
    &                          &  Momask  & 0.276  & 0.441 & 0.552 & \textbf{0.627} & 7.141 & 1.554 & 3.937 \\
    \bottomrule
    \end{NiceTabular}
    \caption{\textbf{Full-body motion generation results}. Models trained and evaluated on the EPFL-Smart-Kitchen-30. FID:Fréchet Inception Distance, DIV:Diversity, MM:Multimodality, MMd:Multimodal distance.}
    \label{tab:text2motion}
\end{table*}

\noindent{\bf Pre-processing, Baselines and Metrics} To achieve robust full-body motion representation, we combine joint locations and angles for the body, hands, and eye gaze, resulting in a 327-dimensional redundant motion representation. We process fine-grained action text with linguistic tags and extract egocentric visual features with CLIP’s text and image encoders~\cite{radford2021learning}. As baselines, we adapt three strong motion generation models, T2M-GPT~\cite{zhang2023generating}, MARDM-SiT~\cite{meng2025rethinking} and MoMask~\cite{guo2024momask}, training them with verb-noun pairs and verb-only prompts (see Supp. Mat.). Like in HumanML3D~\cite{guo2022generating}, we train the quantitative evaluator on EPFL-Smart-Kitchen-30 to measure the R-precision top-1 to top-3 (T1 to T3), multimodal distance (MMd), Fréchet Inception Distance (FID), Diversity (DIV), and Multimodality (MM).

\noindent{\bf Results.} Our qualitative analysis (from MoMask) revealed compelling examples of successfully generated full-body motion sequences that accurately represent natural cooking behaviors (Figure~\ref{fig:motion_generation_examples} and Supp. Mat.). Quantitatively, we found that just like on HumanML3D\cite{guo2022generating}, MoMask \cite{guo2024momask} consistently outperforms T2M-GPT \cite{zhang2023generating} and MARDM-SiT~\cite{meng2025rethinking} in different settings and metrics (Table \ref{tab:text2motion}), possibly due to the hierarchy and quantized tokenization way works better for the high dimensional properties of the whole-body representation. Furthermore, when conditioning on the egocentric view, the model was able to find some clues to minimize the global distribution and thus make the FID score lower.

\section{Conclusion, future work and impact}
\label{sec: conclusion}

We collected 30h of RGB-D video from ten views and 3D pose data as well as hierarchical action annotations in a sensorized, calibrated kitchen. The privacy of all consenting participants was protected; the data will be shared in an anonymized fashion. Our study contributes to understanding  human behavior, encompassing a broad spectrum of object interactions and cognitive processes. The inherent naturalness, multimodality, and precision of the collected motions not only present challenges for computational models but also serve as a robust foundation for the analysis of fine-grained behavioral patterns (Figure~\ref{fig:pose_example_richness}C), strategic decision-making processes, or visual attention mechanisms. Additionally, we are collecting data from older and non-healthy participants (stroke and amputee patients) for future release, aiming to improve treatments for subjects with neurological disorders~\cite{micera2020advanced}. This will also increase the demographic representation of our dataset. Our work complements other recent datasets~\cite{damen2022epickitchen,tanke2024humansinkitchens,grauman2023egoexo4d} that have been collected across multiple kitchens, but lack the multimodality. Future work could leverage more collected modalities (e.g., IMU data and depth) and develop additional benchmarks. 

Overall, we share multi-view, multimodal action understanding, modeling, and video question answering benchmarks to leverage the potential of multi-modality for improving action understanding and to fuel foundation models. This is particularly interesting for emerging multi-modal models~\cite{lucas2022posegpt,feng2024chatpose,Mizrahi20234m}.\\

\noindent{\bf {Acknowledgments:}} We thank members of the Mathis Group for Computational Neuroscience \& AI (EPFL) for their feedback throughout the project. This work was funded by EPFL, Swiss SNF grant (320030-227871), Microsoft Swiss Joint Research Center, and a Boehringer Ingelheim Fonds PhD stipend (H.Q.). We are grateful to the Brain Mind Institute for providing funds for hardware and to the Neuro-X Institute for providing funds for services.

\section*{References}
\bibliography{egbib}

\clearpage
\setcounter{page}{1}
\appendix
\renewcommand{\thesection}{\Alph{section}}

\twocolumn[{\centering \textbf{\Large Appendix to the EPFL-Smart-Kitchen-30 dataset and benchmarks} \par \vspace{20pt}}]

\renewcommand\contentsname{}
\tableofcontents
\counterwithin{figure}{section}
\counterwithin{table}{section}

\clearpage

\section{EPFL-Smart-Kitchen platform}

\subsection{Details on devices and sensors} \label{subsec:sensordetails}

We installed nine Kinect Azure cameras~\cite{kinectazure} inside the kitchen and asked the subjects to wear the HoloLens~2 headset~\cite{ungureanu2020hololens} to record video streams. The RGB streams were recorded at 30 FPS, and the depth streams were recorded at 10 FPS. Meanwhile, we also extracted the head position, hand poses, and eye gaze data at 30 FPS from the HoloLens~2.
Apart from the video streams, we also installed six wired IMU sensors (Adafruit BNO055 sensor) on some of the appliances (fridge/cupboards) and two wireless IMU sensors (Axivity AX3 sensor) on the frequently used tools (knife/spatula). The wired IMU sensors were connected to an Arduino, which also recorded the audio signals from the Kinect Azure cameras and then sent them to those IMU sensors for synchronization. 

The wireless IMU sensors were synchronized based on timestamps recorded during data collection. The wired IMU sensors were recorded at 30 FPS while the wireless IMU sensors were recorded at 100 FPS (Figure \ref{fig:platform} B).


\subsection{Synchronization and calibration of the sensors} \label{subsec:synchronization}

\begin{figure}[h!]
    \centering
    \includegraphics[width = \linewidth]{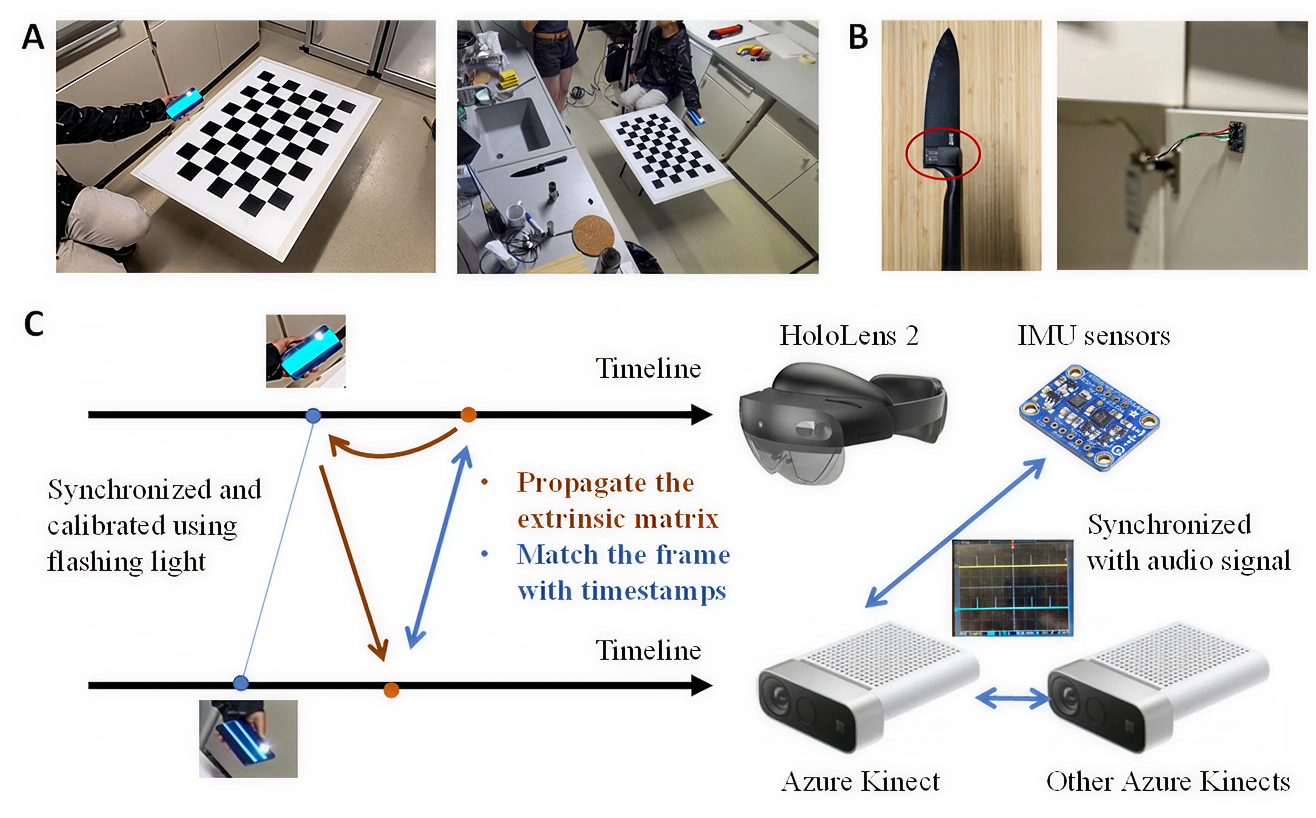}
    \caption{\textbf{Calibration and IMU sensors and synchronization} A). We use an A0-size checkerboard to calibrate egocentric and exocentric cameras at the synchronization timestamp. B). We attach wired IMU sensors to the appliances (fridge/cupboard) inside the kitchen and wireless IMU sensors to the frequently used tools (knife/spatula). C). We synchronize the egocentric and exocentric cameras using the offset of a flashing light and synchronize exocentric cameras and IMU sensors using audio signals.}
    \label{fig:platform}
\end{figure}

We calibrated the egocentric and exocentric cameras using an A0-size checkerboard (Figure \ref{fig:platform} A). Since the egocentric camera is constantly moving during the session, we calibrated only at the calibration timestamp. To calibrate the egocentric camera with exocentric cameras for the whole session, we leveraged the head poses obtained from the HoloLens~2 and did an extrinsic matrix propagation to the calibration timestamp.

As shown in Figure \ref{fig:platform} C, we synchronized the egocentric and exocentric cameras using the offset of a flashing light at the beginning of the session and use the recorded timestamps to match the captured frames. The Kinect Azure cameras can output regular audio signals; we used those signals to synchronize all the Kinect Azure cameras and the IMU sensors. We validated the calibration and synchronization by projecting the regressed 3D body poses to all the views and manually checking if the poses were aligned with the body in all the views at randomly selected timestamps.

\subsection{List of recipes} \label{sec:list_recipes}

Overall, participants were required to cook five times, preparing a different recipe each time, with the exception of ratatouille (section \ref{subsec:ratatouille}), which they made twice. A subset of those sessions is included in the EPFL-Smart-Kitchen-30.

\begin{table*}[h]
    \centering
    \normalsize
    \begin{tabular}{llllll}
    \toprule
    \multicolumn{6}{c}{\textit{Appliances}} \\
    \midrule
    Fridge & Counter & 2$\times$Stoves & Ventilation & Sink & Drawers \\
    Cupboards & Drying rack &  &  &  &  \\
    \midrule
    \multicolumn{6}{c}{\textit{Objects}} \\
    \midrule
    Spatula & Knife & Bottle & 3$\times$Pot & Pot lid & Peeler \\
    Recipe & Sponge & 2 Salad bowls & Cup & Cutting board & Bowl \\
    Colander & Doser glass & Grater & Tissue & Brush & Cleaning gloves \\
    Paper plates & Spoon & Sink Sprayer & Soap & Pasta Spoon & Tissue \\
    Whip & 3$\times$Pan & Towel & Trivet & Fork &  \\
    \midrule
    \multicolumn{6}{c}{\textit{Food}} \\
    \midrule
    Onions & Tomatoes & Avocado & Bell Pepper & Radish & Zucchini \\ 
    Cucumber & Mushrooms & Shallots & Carrots & Eggs & Butter \\
    Surimi & Shrimps & Poultry Broth & Pasta & Noodles & Rice \\
    Vegetable Broth & Olive oil & Frying oil & White wine & Balsamic vinegar & Fish sauce \\
    Basil & Paprika & Tamarind paste & Nutmeg & White sesame & Parsley \\
    Salt & Pepper & Water & Eggplant & Soy sprouts & Thyme \\
    Sesame oil & Lemon & Tofu &  &  &  \\
    \bottomrule
    \end{tabular}
    \caption{Exhaustive list of items present in the EPFL-Smart-Kitchen}
    \label{tab:cooking_list}
\end{table*}

\subsubsection{Omelette and Tomato Salad} 
Tomato salad
\begin{enumerate}
    \item Dice 2 tomatoes 
    \item Mix one spoon of oil with salt, pepper and balsamic vinegar
    \item Pour the dressing on the tomato salad
    \item Stir the salad for at least 2 minutes to allow the tomatoes to soak up the dressing
\end{enumerate}
Omelette
\begin{enumerate}
    \item Beat 3 eggs and season them with salt and pepper
    \item Heat the oil in a pan over a medium-low heat
    \item Pour the eggs into the pan, tilt the pan ever so slightly from one side to another to allow the eggs to swirl and cover the surface of the pan completely
    \item Let the mixture cook for about 20 seconds then scrape a line through the middle with a spatula
    \item Tilt the pan again to allow it to fill back up with the runny egg
    \item Repeat once or twice more until the egg has just set
    \item Fold gently in half with the spatula
\end{enumerate}

\subsubsection{Ratatouille} \label{subsec:ratatouille}
\begin{enumerate}
    \item Put a casserole of salted water to boil.
    \item Heat 2 tablespoons of oil in a pan over medium heat, add the basil stalks and thyme leaves. Cook on a medium heat for 2-3 minutes. During this time, chop the zucchini, eggplant, and mushrooms.
    \item When the water boils, add the pasta for 7-10min depending on the type. Then drain them, drizzle them with olive oil, and keep them aside.
    \item Then add the zucchini, eggplant, and mushrooms (you may need to do this in batches) and fry until golden and softened (approximately 5 minutes). During this time, deseed and finely chop the pepper.
    \item Add the pepper for another 3 minutes. During this time, cut the tomatoes in 4.
    \item Stir in the tomatoes, the balsamic, and a good pinch of sea salt and black pepper. Mix well, breaking up the tomatoes with the back of a spoon. Cook for 5- 7 minutes, stirring the vegetables now and again, until reduced, sticky and sweet. 
    \item Add some basil, finely grate in the lemon zest, and adjust the seasoning if needed. 
\end{enumerate}

\subsubsection{Risotto}
\begin{enumerate}
\item Bring 1.5L of water to a boil in a saucepan and pour in a vegetable stock cube, stir. As soon as it boils, reduce the heat to low and let the broth simmer to keep it warm.
\item Melt the butter in a second pan over medium heat. Add 250 g of risotto rice and cook for about 3 minutes, stirring well, until it becomes translucent. Pour in 10 cl of white wine and cook, stirring frequently, until the rice completely absorbs it.
\item Pour a ladleful of vegetable stock into the pan and continue cooking, stirring occasionally, until the stock has completely evaporated. Once it has evaporated, add another ladleful of broth and wait until it is absorbed again. Repeat several times, adding ladles of broth to the risotto as you go, until the risotto is cooked.
\item Meanwhile, make a vegetable salad. Cut 1-half of cucumber into wedges. Cut the radish into slices, and cut the surimi into slices. Cut the avocado in 2 halves, remove the core, and cut the avocado into thin slices. Peel and grate the carrot and mix everything in a bowl. 
\item Prepare the dressing by mixing the soy sauce and the sesame oil. Pour the dressing over the vegetables and toss for at least 1 minute. Sprinkle with sesame seeds.
\item When the risotto is cooked, add a spoonful of butter and 3 spoonfuls of grated Parmesan cheese. Add salt and pepper to taste and stir well.

\end{enumerate}

\subsection{Pad Thai}
\begin{enumerate}
\item Dip the noodles in hot water to soften them
\item Dissolve a portion of tamarind paste in 100 ml of hot water, stir, and filter the
mix to obtain tamarind juice.
\item Cut the tofu in cubes, the shallot, and the onion in slices.
\item Heat up the frying oil in a pan and cook the tofu until it becomes golden,
reserve for later
\item Add the drained noodles and cook them with the juice in the pan, add water
if the noodles absorb too much.
\item Add the egg to the pan next to the noodles and cook the egg before mixing
the noodles, you should obtain separate white and yellow parts.
\item Add the tofu, the peanuts, and the dried shrimp
\item Remove from the heat and add the soy sprouts, the onion, and the shallot,
serve with a lemon slice.
\end{enumerate}

\subsection{List of kitchen items}

Table \ref{tab:cooking_list} lists all items present and used in the kitchen by the participants.

\begin{figure*}[ht!]
    \centering
    \includegraphics[width=\linewidth]{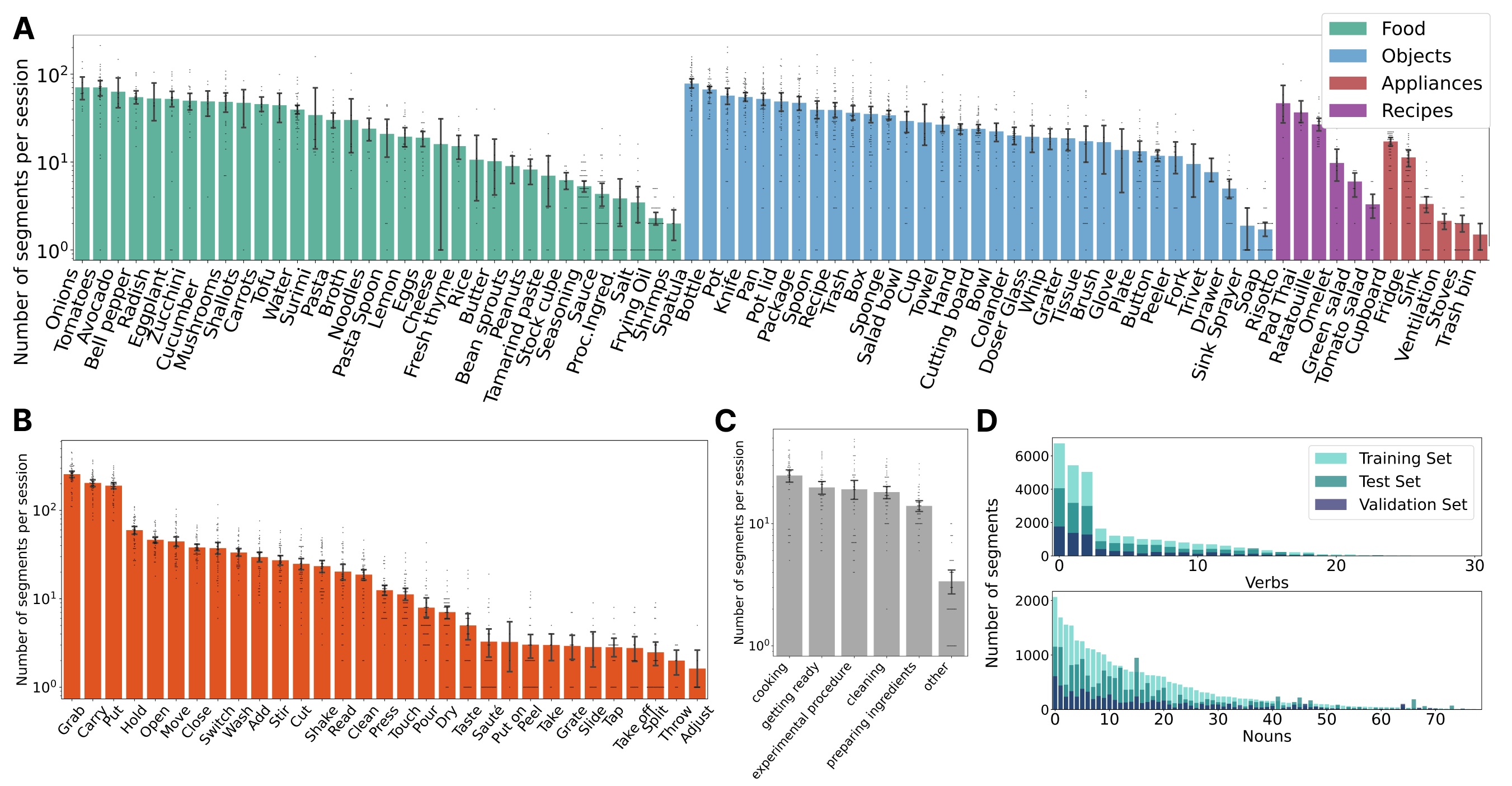}
    \caption{\textbf{Number of action segments}. (A) Nouns, (B) Verbs, (C) Activity (D) Number of segments per split for nouns (top) and verbs (bottom). N = 49.}
    \label{fig:supp_distribution}
\end{figure*}

\section{Details on action annotations}

\subsection{Action distributions}

Our EPFL-Smart-Kitchen-30 provides action segments of various lengths with a long-tail distribution. To support this, we show the duration of verbs and nouns in Figure \ref{fig:supp_distribution}.

\subsection{Verbs descriptions}
\label{sec:verb_description}
Each verb is provided with a detailed description to ensure specific usage and avoid confusion during annotations. Below is an exhaustive list of the verb descriptions along with general annotation rules. We note that these instructions could be used for NLP-based systems in the future. 

\begin{itemize}
\item\textit{Open/Close}: Opening/Closing objects such as Pot, Package, Butter, Tamarind paste, Water, Oil, Sauce, or appliances/objects such as Fridge, Cupboard, Drawer.  

\item\textit{Cut}: Operation of Cut an object or food when Carrying a Knife. Exceptions: The action of opening a package with a Knife should be defined as Open. 

\item\textit{Wash}: Washing an object or food requires the use of a certain amount of Water and Soap and should happen in the sink area. To be contrasted with Clean which is reserved for areas apart from the Sink. 

\item\textit{Clean}: Cleaning an object/food requires the use of a Tissue, Towel, sponge, or Hand. It can happen in any area apart from the Sink area. It could involve motions such as wiping or removing food from the Knife in a Cut action. 

\item\textit{Stir}: Motion of stirring which can happen in Salad Bowls, Pan, Pot, etc.  

\item\textit{Wait}: This action is not related to any noun. Wait action should last for at least 3 seconds, otherwise there is no need to label this segment. 

\item\textit{Put}: Action of setting down an object.  

\item\textit{Split}: Action of separating Processed Ingredients in two parts. To be contrasted with Cut which mostly happens with which is also splitting but with a Knife. Cut must be prioritized over Split if in hesitation. 

\item\textit{Touch}: The action of barely touching and getting in contact with an object and having absolutely no consequences, can be involuntary. To be contrasted with Tap (repetitive movement) and Press (voluntary and has a certain degree of consequences), with more force). Touch does not need to be annotated during other behaviors that involve touch (e.g. Stir, Tap, ..) 

\item\textit{Peel}: Removing the skin of a specific food. Can happen with a Peeler, a Knife, or directly with Hands. 

\item\textit{Pour}: The action of setting down transferring a liquid (Water, Oil, Sauce, Broth)  from one recipient into another recipient (such as from a Bottle into a Pan, or from a Bowl into the Sink). Note that Tamarind paste is also considered as liquid and follows the Pour rule. 

\item\textit{Sauté}: Sautéing represents the movement where the cook shakes the Pan instead of Stir. It is only used together with Pan. 

\item\textit{Taste}: The action of taking putting food in the mouth. The action should be annotated from the moment the food reaches the mouth of the participant and should be following Grab and Carry. 

\item\textit{Tap}: Tap with the Spatula, Spoon, or Knife often to remove the stuck residues. 

\item\textit{Add}: Add a set of ingredients to another set of ingredients. Add has a higher priority than Put. For example, we prefer Grab/Carry/Add Pasta instead of Grab/Carry/Put Pasta when the subject wants to cook the pasta and add it to the Pan. 

\item\textit{Throw}: Defined when the object is thrown in the air.  

\item\textit{Carry}: Defined when the interacting object is lifted. Intermediary action defined in between Grab and Put/Add.  

\item\textit{Hold}: Defined when the person tries to stabilize the object with almost no placement. Hold must be annotated also during Cut if the participant Holds with the contralateral hand.  

\item\textit{Move}: Move is an action that involves the displacement of an object without lifting it (slide). Move is not followed by Put. 

\item\textit{Dry}: Action happening after Wash. Drying requires the use of a Towel or Tissue. Involve the action of wiping on an object. To be contrasted with Shake when performed without Towel or Tissue. 

\item\textit{Grab}: Grab an object, must typically be followed by Carry (lifted) and eventually Put or Hold (not lifted). Grab is defined from the moment when the reaching movement starts and ends at the hand-object contact instant. Note that some sequences can also have Grab followed by Hold then followed by Carry. 

\item\textit{Read}: Applies to Recipe but also to any object with associated text such as Package or Bottle. 

\item\textit{Press}: Typically Press Button when the participant sets the Stoves or Press Hands when the participant Open/Close the HoloLens menu by pressing their wrist. To be contrasted with Tap (repetitive movement with an object) and Touch (No consequences of the action). Press Button can be defined once for repetitive instances of the actions as long as the arm of the participant does not retract but must not intersect with Slide Button (For the slide button in the middle of the stoves) 

\item\textit{Slide}: Correspond to the sliding of a hand or an object on a surface or other object. The participants have the possibility to Slide the Button of the stoves. 

\item\textit{Shake}: Move an object from side to side with a forceful, jerky movement. Should be used instead of Dry when relevant. 

\item\textit{Squat}: Knees bent and one's heels close to or touching one's buttocks or the back of one's thighs. The Squat action is not associated with any noun. Often happens in front of the fridge. It represents the whole state rather than only the movement from the start of the crouching to the end of the standing-up action. 

\item\textit{Switch}: Switch an object from one hand to another, annotate when both hands are in contact with the object. 

\item\textit{Grate}: It happens when the subject uses the Grater (object) to process the food (e.g., carrots or cheese). 

\item\textit{Put on/ Take off}: They can be used exclusively with Glove or other clothes 

\item\textit{Take(with tool)}: to describe the action when the object is lifted by a tool. It is used to describe a status, is not necessarily followed by Put. 
\end{itemize}

\subsection{Clarification of special Food/Objects}
\label{sec:noun_description}

\begin{itemize}
\item Tamarind Paste is only used in the Pad Thaï recipe as a paste mixed with water. After dilution please refer to it as Sauce 

\item Broth is only used in the Risotto recipe after the stock cube is Mixed with the water, the participant typically Add Broth to the Rice 

\item To avoid confusion, Frying Oil should only be used when Frying Oil is added to the pan. Otherwise, it is considered a Sauce. 

\item \textit{Sauce}: Any cooking liquid including oils, vinegar, wine and liquid mixes. Adding seasoning to a sauce does change its nature. Exception: Frying oil 

\item \textit{Salt}: To avoid confusion, Salt must be only used when adding salt to the pasta water. Otherwise it is considered as Seasoning. 

\item Pasta is only present in the ratatouille (wheat pasta) 

\item Noodles are used in Pad Thaï (rice noodles). 

\item Seasoning involves any herbs, salts or spices added with a shaking motion. Exception: Salt when added to the water of the pasta. 

\item \textit{Bowl and salad bowl}: The EPFL-Smart-Kitchen possesses 2 metallic salad bowls of different sizes and metallic. Any other spherical-shaped container can be considered as a Bowl. It can involve the small yellow bowl and eventually the lunch boxes. 

\item Processed ingredients are normally referred to as “uncooked food”. If the food mix is cooked, then try to call it with the corresponding recipe. 

\item Specific rules for annotation using a limited set of words: 

\item Actions can occur simultaneously 

\item Food is defined by their specific names only if they are targeted by the action.  

\item When food is processed but not yet cooked please use processed ingredients. By definition, this does not apply to Salads (as it does not require cooking). 

\item If all ingredients are added to the mix and the participant manipulates the mix, please use the name of the corresponding recipe instead. For instance,  in the Ratatouille recipe, when the tomatoes, zucchini, and eggplant are mixed together, then it is labeled as ratatouille. 

\item An object should be defined as Trash as late as possible before going to the Trash bin. The object should keep its name (if known) as long as possible, the last action should still include Trash in its name. (example Grab Zucchini, Put Trash) 

\item Please be as precise as possible on the object interaction, for instance, the participant will never Carry a seasoning but a bottle - they Carry a Bottle. However, they may add the Seasoning. 

\item All the actions should last for at least 5 frames. 

\end{itemize}

\begin{figure*}[h]
    \centering
    \includegraphics[width = \linewidth]{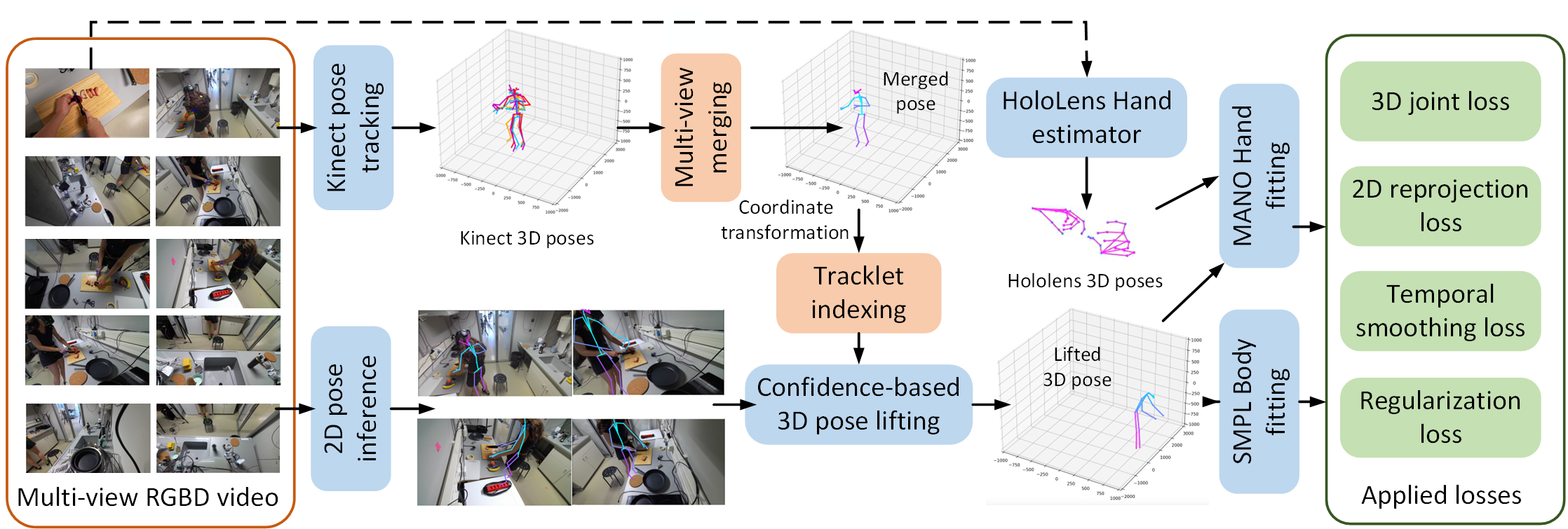}
    \caption{\textbf{Body/Hand motion estimation pipeline.} We effectively use multi-view and multi-modal information to regress accurate 3D body and hand meshes across the whole video sequence.}
    \label{fig:ESK_3Dpose}
\end{figure*}

\subsection{Annotator training}

Our research utilized the services of a third-party commercial annotation company specialized in data labeling. 

To ensure high-quality labels, we trained annotators with two mini-batches of data. Specifically, we first asked the annotators to review our annotation requirements (see section \ref{sec:verb_description} and section \ref{sec:noun_description}). After that, we gave them the mini-batch of data for initial annotation. Meanwhile, two authors of this paper manually annotated the mini-batch of data. The annotated actions were merged by manual check to serve as the gold standard, which was given to the annotators after they finished their own annotations. After two rounds of training, and when annotations from all the annotators could achieve an F1 classification score larger than 0.9, the annotators started to collect ground truth annotations.

\subsection{Validation of action annotations}
We perform two ways of annotation validation jointly to ensure the annotation quality. Firstly, we asked additional annotators to randomly check the annotations of certain clips. Videos with unsatisfactory clip annotations will be returned to the annotators for second-round annotation. Secondly, we applied a rule-based check for all the annotations. For example, we check if the `Cut` action segments are inside the `Carry Knife` action segments. Unsatisfactory action segments were returned to the annotators for second-round annotation as well.

\section{Automatic 3D body/hand motion estimation} \label{sec:3dpose}

To efficiently and accurately capture a large volume of hand and body motions from the captured videos, we propose a body/hand motion estimation method that can effectively use multi-view and multi-modal information (Figure \ref{fig:ESK_3Dpose}). Specifically, the body/hand motion estimation method can be divided into four steps: 2D keypoint estimation, identity tracking, 3D keypoint lifting, and 3D mesh fitting. We will detail each step in the following sections. Furthermore, we also compare the bone length variance between our estimated hand poses with the HoloLens~2 estimated hand poses to show the effectiveness of our estimated poses.

\subsection{2D keypoint estimation}

To balance the trade-off between performance and efficiency, we choose to use RTMPose~\cite{jiang2023rtmpose} for 2D pose estimation; the model is available in DeepLabCut v3~\cite{mathis2018deeplabcut}. Specifically, we use the RTMPose-x model trained on eight pose estimation datasets to extract body poses~\cite{jiang2023rtmpose}. Due to the small size of the hand regions in the video, the existing hand detectors often struggle with hand-missing detection and, due to occlusion or interactions, are prone to errors. We choose to use the whole-body model (RTMW-x) trained on 14 public datasets to extract hand poses, because it relies on semantic relations to effectively find the hand regions.

\begin{figure*}[h]
    \centering
    \includegraphics[width = \linewidth]{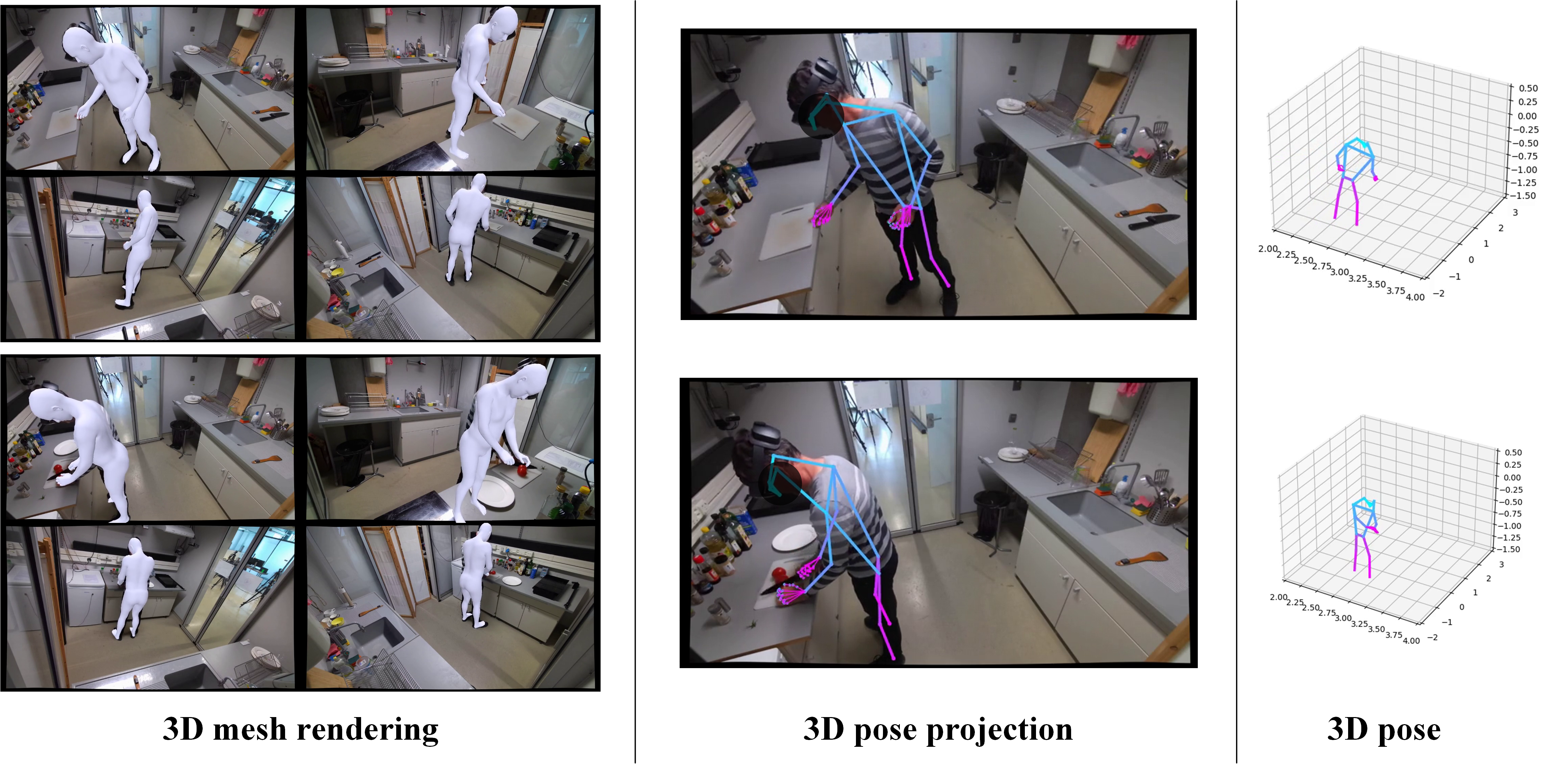}
    \caption{\textbf{3D pose examples.} Examples of the estimated Body/Hand meshes (left), reprojected 3D pose (center) and 3D pose (right).}
    \label{fig:3D_visualization}
\end{figure*}

\subsection{Identity tracking}
The experimentalists occasionally appear in the data collection area to assist the subjects. When this occurs, RTMPose \cite{jiang2023rtmpose} can misalign detection identities for the detections in previous frames. Therefore, before grouping the 2D pose estimation from each camera for 3D pose estimation, it is necessary to identify the 2D pose of the subject in each view across the entire video. To achieve this, we rely on the Kinect body tracking SDK \cite{kinectbody} to determine the correct identity of the subject. It estimates the 3D body pose and the tracking identity using the depth images from each camera view. We first transform the 3D body poses into the world coordinate system and merge them to obtain a robust 3D body pose estimation. After that, we consider the dominant identity detected in the video as the subject and project that 3D body pose into each camera view. We use the projected 2D body pose to identify the correct 2D poses estimated by RTMPose \cite{jiang2023rtmpose}. We do not use the merged 3D body pose intermediate results as the final 3D pose results as they are quite jittery.

\subsection{Confidence-based 3D keypoint lifting}
With identity information, we group the 2D poses of the subject from all camera views for 3D pose lifting. We perform singular value decomposition (SVD) using both the 2D pose results and the corresponding confidence scores from all camera views. Pose joints with scores lower than a certain threshold (0.1) are excluded from subsequent 3D mesh fitting.

\subsection{3D mesh fitting}

We adopted the multi-view mesh fitting implementation of EasyMocap \cite{easymocap} to estimate the body and hand meshes of our data. Specifically, we utilized lifted 3D poses and 2D poses estimated by RTMPose \cite{jiang2023rtmpose} to fit the SMPL parametric model \cite{SMPL-X:2019}. We modified the original joint regressor matrix to adapt it to the COCO \cite{lin2014microsoft} joint definition. 3D joint loss, 2D reprojection loss, temporal smoothing loss, and regularization loss are minimized to optimize the pose, shape, and global transformation parameters of SMPL \cite{SMPL-X:2019}. Meanwhile, the 3D hand poses obtained by the HoloLens~2 hand tracking toolkit \cite{holohand} are estimated from the egocentric view depth image, which contains hand contact information that RGB views may lack. Therefore, we also use the relative 3D hand poses from HoloLens~2 to minimize an additional 3D joint loss for fitting the MANO parametric model \cite{MANO:SIGGRAPHASIA:2017}. Finally, we use the fitted SMPL and MANO models to obtain the final 3D body and hand poses (Figure \ref{fig:3D_visualization}).We discard frames where the fitted pose significantly differs in Euclidean distance from the lifted pose, as this discrepancy is likely caused by artifacts in the fitting algorithm.

\subsection{Assessment of Pose Estimation Quality}

Creating 3D ground truth is notoriously difficult. We measure the quality of the 3D pose estimation pipeline by manually annotating 4,947 frames with 2D keypoints for a fixed, random selection of 14 keypoints from all exocentric views. The frames were extracted using DeepLabCut's frame extraction pipeline; specifically, we sampled distinct visual frames via k-means~\cite{mathis2018deeplabcut}. 

\begin{figure*}[h]
    \centering
    \includegraphics[width = \linewidth]{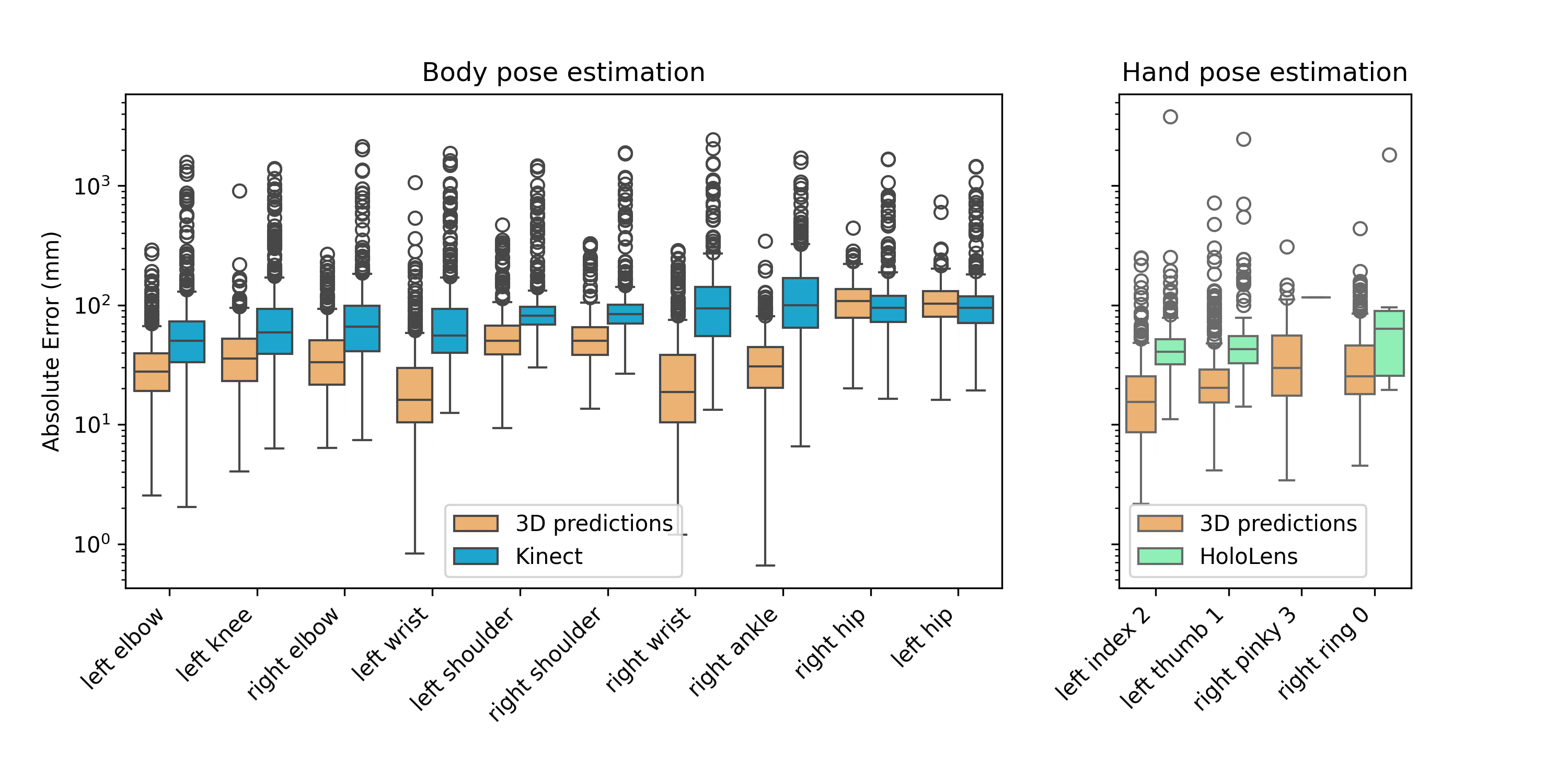}
    \caption{Absolute error with triangulated ground truth comparing our 3D pose estimation (N=469) to 3D body prediction from the Kinect Azure cameras (N=469) and HoloLens~2 hand pose estimation (N=157). 
    3D hand pose estimation}
    \label{fig:pose_validation_kinect_hololens}
\end{figure*}

We then compare the mean euclidean distance  between our predicted poses and the triangulated annotated ground truth data. Our pose estimation shows an average error of 52.75 mm ± 53.26 mm, where 31.73 mm ± 43.06 mm is from the hands and 57.69 mm ± 54.22 mm is from the body. These findings are based on 469 triangulated poses (Figure \ref{fig:pose_validation_kinect_hololens}). Our pose estimation provides results closer to the ground truth compared to the original merged 3D body pose estimations from Kinect Azure cameras and HoloLens~2 3D hand poses (Figure \ref{fig:pose_validation_kinect_hololens}). The Kinect Azure 3D pose estimation has an error of 118.37 mm ± 177.76 mm (for the body), while the HoloLens~2 estimation has an error of 86.20 mm ± 297.28 mm, comparing only frames with visible hands in an egocentric view.

\section{Protocols} \label{sec:ethics} 

\subsection{Ethics protocol} 

The study was approved by the Swiss Ethics Committees on research involving humans (Project ID: 2022-00493 ). Following the approved protocol, the participants sign a research agreement before they start participating. Here, we share a summary for the community (All content in italics is copied from the original ethics protocol):

\begin{itemize}
    \item \textbf{Motivation of this study} : \\
    \textit{The goal of this project is to build a database of subjects using a kitchen in a “natural world” scenario. We are collecting data on healthy subjects and on subjects with motor deficits. These data will help us to build new metrics (standardized evaluations) using machine learning to monitor patient's progress during their rehabilitation process or during the use of assistance devices.}
    \item \textbf{Procedure of a typical cooking session :} \\
    \textit{If you accept to participate in our project, you will use a kitchen equipped with wall-fixed cameras, one head-fixed camera that will record your arms and track your eyes, and cooking utensils equipped with movement and force sensors. You will also answer some surveys. You will be allowed to use the kitchen at your own convenience, taking into consideration the following conditions: \\
    (1) Preparation of the meal \\
    (2) Cooking (for instance cooking pasta) and preparation of drinks (for instance coffee, tea). \\
    (3) Cleaning of the cooking utensils for the next participant.} \\
    \item \textbf{Benefits and risks or constraints related to the participation in this study }: \\
    \textit{By participating in this study, you will only be exposed to minor inconveniences, such as the recording of your movements and carrying the head-mounted camera. There are no particular risks related to the recording except the two cited in a normal kitchen (hot water, use of a knife, etc …)}
    \item \textbf{Anonymization of the collected data }: \\
    \textit{If your face or any distinctive feature (such as tattoos) appears on one of the videos, they will automatically be detected and blurred at the end of the recording session. We will only conserve the videos the edited videos with your face, or any other distinctive sign blurred, and delete the original data. \\
    Your data will be accessible to a limited number of researchers working on this study. These data will be pseudoanonymized (which means that your identification information (name, birthday) will be replaced by a code). Your identity key will be secured. \\
    Your data will be useful for the creation of the dataset. At this point, your data will be anonymized (which means that your personal data will be aggregated and protected, preventing anyone from identifying you. Your name will never appear on the internet or on a publication.}
    \item \textbf{Optional nature of participation and obligations}
    \textit{Your participation is completely free. If you choose not to participate or if you choose to participate and reconsider your decision during the course of the project, you will not have to justify yourself. This decision will not have a negative impact on the rest of your medical care. If you choose to participate in this research project, you will be required to follow the instructions and fulfill the requirements of the research protocol.}
    
\end{itemize}

\subsection{Data collection protocol} 

The data collection process of EPFL-Smart-Kitchen-30 was reviewed and approved by the Swiss Ethics Committee. Data capture occurs only in the EPFL-Smart-Kitchen at Campus Biotech, Geneva CH. All subjects underwent a MoCA \cite{nasreddine2005montreal} test before starting the first cooking session to assess for mild cognitive impairment; subjects with a score lower than 25 were excluded from the study. 

Each subject participated in five cooking sessions ranging from 30 minutes to 1 hour throughout the study. They were presented with four recipes in the following order: Omelet and Tomato salad, Ratatouille, Risotto and green salad, Ratatouille, Pad Thai (see \ref{sec:list_recipes} for the content of the recipes). Ratatouille was cooked twice to study subjects' adaptation. At the start of each session, the participant followed the calibration protocol of the HoloLens~2 headset for eye tracking. 

At the beginning of each session, subjects were given the following instructions:

\begin{itemize}
    \item Prioritize using the knife and spatula because they are equipped with some of the IMU sensors.
    \item Keep their hand movements within the visual field of the HoloLens~2 camera.
\end{itemize}

The session started after the calibration and synchronization frames of all 10 cameras were recorded. During each session, subjects could ask for clarifications on the kitchen appliances (particularly the stoves) and cooking methods.

As stated in the ethical protocol, we blurred the participants' faces for all the shared data.

\section{Compute resources}

Table \ref{tab:ComputRessources} outlines the details of the computer resources used to collect the video data, estimate the 3D body/hand motions, and train each benchmark model.

\begin{table}[h!]
    \small
    \centering
     \renewcommand{\arraystretch}{1}
     \begin{NiceTabular}{Wl{35pt} Wl{190pt}}
         \toprule
         \multicolumn{2}{c}{Data collection} \\
         \midrule
         Computers & \fivelines{We used four computers during data collection to}{avoid frame losses due to buffer overload. Three}{computers collected data from three Kinect Azure}{cameras while the last computer collected exclusively}{the HoloLens~2 videos.}\\
         RAM & 3$\times$ 32GB\\
         \midrule
         \multicolumn{2}{c}{3D body/hand motion estimation}\\
         \midrule
         GPUs & 1 $\times$ RTX 3080 Ti \\
         RAM & 1 $\times$ 32 GB \\
         \midrule
         \multicolumn{2}{c}{Question-answering benchmark} \\
         \midrule
         GPUs & 4 $\times$ A100 \\
         RAM  & 1 $\times$ 64 GB \\
         \midrule
         \multicolumn{2}{c}{Action recognition benchmark} \\
         \midrule
         GPUs & 2 $\times$ A100 \\
         RAM & 1 $\times$ 64 GB \\
         \midrule
         \multicolumn{2}{c}{Action segmentation benchmark} \\
         \midrule
         GPUs & \threelines{GeForce RTX 3090 and A100.To run experiments in}{parallel we used computers with different GPU settings}{to train models on the action segmentation task} \\
         RAM & 64GB \\
         \bottomrule
    \end{NiceTabular}
    \caption{Details on computational resources used for data collection, 3D pose estimation and benchmarking.}
    \label{tab:ComputRessources}
\end{table}

\section{Supplementary material for benchmarks}

\subsection{Details on Lemonade}

\subsubsection{Comparison with other benchmarks}

Lemonade is well situated compared to other benchmarks for assessing human motion. Lemonade has the particularity of using ground truth action annotations and pose estimation for designing QA pairs (Table \ref{tab:qa_dataset_comparison}).

\begin{table}[h]
    \renewcommand{\arraystretch}{1.}
    \centering
    \scriptsize
    \setlength{\tabcolsep}{3pt}
    \begin{NiceTabular}{lccccc}[code-before = \rowcolor{lightgray}{9}]
    \toprule
    \twolines{Question answering}{video datasets} & \# Videos & \twolines{\# Closed}{-ended QA}  & \twolines{Ego}{-centric} & \twolines{From action}{annotations} & \twolines{From pose}{estimation} \\
    \midrule
    ActivityNet-QA~\cite{yu2019activitynet}    & 5,800  & 58,000 & \xmark & \cmark & \xmark\\
    MVBench~\cite{li2024mvbench}               & 4,000  & 4,000  & \xmark & \xmark & \xmark\\
    TVBench~\cite{cores2024tvbench}            & 2,525  & 2,525  & \xmark & \xmark & \xmark\\
    MLVU~\cite{zhou2024mlvu}                   & 1,730  & 3,102  & \cmark & \xmark & \xmark\\
    MotionBench~\cite{hong2025motionbench}     & 5,385  & 8,052  & \xmark & \xmark & \xmark\\
    EgoSchema~\cite{mangalam2023egoschema}     & 5,031  & 5,031  & \cmark & \xmark & \xmark\\
    EgoTaskQA~\cite{koyejo2022egotaskqa}       & 2,315  & 40,322 & \cmark & \cmark & \xmark\\
    \textbf{Lemonade (ours)}                   & 36,521 & 36,521 & \cmark & \cmark & \cmark\\
    \bottomrule
    \end{NiceTabular}
    \caption{Comparison with other QA benchmarks related to motion and movement.}
    \label{tab:qa_dataset_comparison}
\end{table}

\subsubsection{Question distributions per category}

Questions are selected and filtered from a pool of 5,116,194 questions, automatically generated through a combination of clips, question types, and answer ranges. The dataset was sampled and filtered based on frame visibility, presence of hands in specific questions, and manual curation. Lemonade includes 18,857 questions in the Behavior Understanding category, with 9,518 in the Perception subcategory and 9,339 in the Reasoning subcategory. There are 8,201 questions in the Long-term Understanding category, with 2,065 in the Session Properties subcategory and 6,136 in the Summarization subcategory. Lastly, 9,463 questions belong to the Motion and Biomechanics category, with 5,916 in the Kinematics subcategory and 3,547 in the Physical Attribute subcategory. Table \ref{tab:lemonade_questions} shows an exhaustive list of the question types. 

\subsubsection{Intuition behind categories and subcategories}
Behavior understanding questions centered around annotated actions and activities.
\begin{itemize}
    \item \textbf{Perception}: Questions directly asking which behavior are visible in the clip. This loosely compares to traditional tasks on action recognition or action segmentation.
    \item \textbf{Reasoning}: Questions asking about behaviors not visible in the frames (performed before or after). Context information should feed a reasoning method to answer these questions.
\end{itemize}

Long-term understanding questions use long-duration clips as input.
\begin{itemize}
    \item \textbf{Session properties}: Questions about session duration or participants age. These questions evaluate if general information can be inferred from long clips.
    \item \textbf{Summarization}: Questions on action sequences or number of instances given actions are performed in a session.
\end{itemize}

Motion and Biomechanics questions take pose estimation as reference information.
\begin{itemize}
    \item \textbf{Physical attributes}: Question related to static pose information.
    \item \textbf{Kinematics}: Question related to speed and motion.
\end{itemize}

\subsubsection{List of questions}
Table \ref{tab:lemonade_questions} lists all question templates and the number of questions per type.

\begin{table}[h]
    
    \centering
    \scriptsize
    \begin{tabular}{clc}
        \toprule
        QID & Question examples & \# Questions \\
        \midrule
        \multicolumn{2}{c}{Behavior understanding --- Perception}\\
        \midrule
        0 &"What action am I doing ?" & 1003 \\
        1 &"What activity am I doing ?" & 1025 \\
        2 &"How many actions am I doing ?" & 1066 \\
        3 &"What am I doing with the \textit{knife} ?" & 1017 \\
        4 &"For how long am I \textit{holding} the \textit{pan} ?" & 987 \\
        5 &"For how long am I \textit{cleaning} ?" & 1054 \\
        13&"What am I \textit{pouring} ?" & 1027 \\
        14&\twolinesleft{"How much time passes from}{\textit{cutting} the \textit{zucchini} to \textit{grabbing} the {zucchini}?"} & 1037 \\
        15&\twolinesleft{"At what moment does \textit{carrying} the}{ {tomatoes} \textit{starts} in the clip?"} & 1031 \\
        \midrule
        \multicolumn{2}{c}{Behavior understanding --- Reasoning} \\
        \midrule
        6 &\twolinesleft{"I am currently \textit{carrying} }{the \textit{tomatoes}, what will be the next action(s) ?"} & 1030\\
        7 &"I am currently \textit{holding}, what \textit{was the previous action} ?" & 1021\\
        8 &\twolinesleft{"I am currently \textit{cooking at the stoves},}{what \textit{will my next activity be} ?"} & 1234\\
        9 &"What were my previous \textit{3} actions ?" & 2096\\
        10&"What will be my next \textit{2} actions ?" & 2184\\
        11&\twolinesleft{"I am currently \textit{grabbing} the}{\textit{zucchini}, what \textit{was my previous activity} ?"} & 1008\\  
        12&"I am currently \textit{pouring}, what \textit{is my current activity} ?" & 1037\\
        \midrule
        \multicolumn{2}{c}{Long-term understanding --- Sessions properties} \\
        \midrule
        16&"What is my age ?" & 1025\\
        17&"What recipe am I cooking ?" & 1040\\
        \midrule
        \multicolumn{2}{c}{Long-term understanding --- Summarization} \\
        \midrule
        18&How many times am I \textit{grabbing} the {peeler} in this session ?" & 1000\\
        19&"How many times am I \textit{tasting} in this session ?"& 988\\
        20&"For how long am I cooking ?"& 1004\\
        21&"What is the correct sequence of action ?"& 2150\\
        22&"What was the longest action in this session ?" & 994\\
        \midrule
        \multicolumn{2}{c}{Kinematic and Biomechanics --- Physical attributes} \\
        \midrule
        23&"What is the average height of my eyes in this clip ?" & 1177\\
        24&"What is the average shape of my \textit{right} hand in this clip ?" & 1200\\
        25&"What is my average trunk bending angle in the clip?" & 1170\\
        \midrule
        \multicolumn{2}{c}{Kinematic and Biomechanics --- Kinematics} \\
        \midrule
        26&\twolinesleft{"The clip lasts \textit{1.43}s, What is the}{ average speed of my \textit{left hand} in this clip ?"} & 1187 \\
        27&\twolinesleft{"The clip lasts \textit{1.73}s, At what speed}{ am I reaching for the \textit{box}?"} & 1180 \\
        28&\twolinesleft{"The clip lasts \textit{1.6}s, At what speed am}{ I putting the \textit{bowl} down?"} & 1186 \\
        29&\twolinesleft{"What is the \textit{minimum} angle between my \textit{left shoulder},}{my \textit{right shoulder} and my \textit{right elbow} in this clip ?"} & 1163 \\
        30& "What is the \textit{maximum} distance between my hands?" & 1200 \\
        \bottomrule

    \end{tabular}
    \caption{Base questions with their identifiers and number of samples. In italic, elements subject to change.}
    \label{tab:lemonade_questions}
\end{table}

\subsubsection{Question engineering}

\noindent\textbf{Q0: What action am I doing ?}\\
\noindent\textbf{Rational.} This question challenges VLMs to perceive the action present in the clip.\\
\noindent\textbf{Design.} Action segments were sampled and used as reference clips. Answers are actions (verb + noun) or verbs randomly sampled from the action or verb list.

\noindent\textbf{Q1: What activity am I doing ?}\\
\noindent\textbf{Rational.} This question challenges VLMs to perceive the activity present in the clip (coarse-grained action)\\
\noindent\textbf{Design.} Activity segments were sampled and used as reference clips. Answers are actions (verb + noun) or verbs randomly sampled from the action or verb list.

\noindent\textbf{ Q2: How many actions am I doing ? }\\
\noindent\textbf{Rational.} Based on the precise definition we give of actions, the number of actions present in a clip is a fixed number. This question challenges VLMs to count the number of actions that happen in a short clip.\\
\noindent\textbf{Design.} For clips ranging from Xs to Ys, we count the number of segments that start within the clip. Answers are randomly sampled from  to 5.

\noindent\textbf{ Q3: What am I doing with the [NOUN]?}\\
\noindent\textbf{Rational.} Prediction of verbs based on nouns.\\
\noindent\textbf{Design.} Action segments were sampled and used as reference clips. We extract the noun from the action for the question and the verb for the answers. Answers are verbs randomly sampled from the verb list.

\noindent\textbf{ Q4: For how long am I [VERB] the [NOUN] ?}\\
\noindent\textbf{Rational.} Estimating timings from context is difficult both for VLMs and humans. This question challenges models to estimate timings based on context.\\
\noindent\textbf{Design.} Action segments were sampled and used as a reference clip. We extract the correct answer from the clip length. Other answers are samples from windows with sizes corresponding to 90\%, 50\%, and 20\% difference with the correct answers for difficulties in easy, medium, and hard, respectively.

\noindent\textbf{ Q5: For how long am I [VERB] ?}\\
\noindent\textbf{Rational.} Similar to Q4 but based on verbs only.\\
\noindent\textbf{Design.} Similar to Q4 but using clips based on verb segments.

\noindent\textbf{ Q6: I am currently [VERB] the [NOUN], what [BE$|$past,future] [NEXT/PREVIOUS] action ?}\\
\noindent\textbf{Rational.} This question challenges models to reason about actions that would happen before and after the visible clip. This requires a strong understanding of the context and the participants' behaviors. Note that predicting previous actions is more difficult both for humans and models.\\
\noindent\textbf{Design.} Action segments were sampled and used as a reference. The correct answer corresponds to the action that starts after/before the start of the current action. Other answers are randomly sampled from the action list.

\noindent\textbf{ Q7: I am currently [VERB], what [BE$|$past,future] [NEXT/PREVIOUS] action ?}\\
\noindent\textbf{Rational.} Similar to Q6 but based on verbs only.\\
\noindent\textbf{Design.} Similar to Q6 but using clip and answers based on verbs.

\noindent\textbf{ Q8: I am currently [ACTIVITY], what [BE$|$past,future] [NEXT/PREVIOUS] activity?}\\
\noindent\textbf{Rational.} Similar to Q6 and Q7 but based on activities only.\\
\noindent\textbf{Design.} Similar to Q6 and Q7  but using clip and answers based on activity.

\noindent\textbf{ Q9: What were my previous [NUMBER] actions ?}\\
\noindent\textbf{Rational.} The rationale is similar to Q5, Q6, and Q7, with an additional difficulty being the possible answers also sampled from previous actions. The models are required to rank the probabilities that a set of actions is happening right before rather than categorizing them.\\
\noindent\textbf{Design.} The window size is fixed at 50 frames. We extract 2,3,4 (easy, medium, hard) actions from the window before the current clip. Answers prioritize the actions that are present in the video but happening before the extracted actions. The other actions are randomly sampled from the action list.

\noindent\textbf{ Q10: What will be my next [NUMBER] actions ?}\\
\noindent\textbf{Rational.} Similar to Q9 but using sets of actions performed after the clip.\\
\noindent\textbf{Design.} Same as Q9 but based on the window happening after the current clip.

\noindent\textbf{ Q11: I am currently [VERB] the [NOUN], what [BE$|$past,present,future] my [NEXT/CURRENT/PREVIOUS] activity ?} \hfill\\
\noindent\textbf{Rational.} Prediction of the activity based on the current action.\\
\noindent\textbf{Design.} Similar to Q6,Q7, and Q8 but clips based on actions and answers based on activities.

\noindent\textbf{ Q12: I am currently [VERB], what [BE$|$past,present,future] my [NEXT/CURRENT/PREVIOUS] activity ?}\hfill\\
\noindent\textbf{Rational.} Prediction of activity based on current action (verb).\\
\noindent\textbf{Design.} Similar to Q11 but using clip based on verbs instead.

\noindent\textbf{ Q13: What am I [VERB]?}
\noindent\textbf{Rational.} Prediction of the noun based on the current action (verb).\\
\noindent\textbf{Design.} Similar to Q3, clips are sampled from verb segments, and answers are nouns randomly sampled from the noun list.\\

\noindent\textbf{ Q14: How much time passes from [VERB] the [NOUN] to [VERB] the [NOUN] ?}\\
\noindent\textbf{Rational.} Estimating timings is difficult for VLMs but also for humans. This question is designed to test the ability of the model to estimate time between two actions. The question is designed to be difficult for humans as well.\\
\noindent\textbf{Design.} The window size is fixed at 200 frames. For each action ending, we look for the previous action starting within the window. The correct answer is the time between the two actions. There is no minimum gap (the answer can be 0). The difficulty is set by the range from which the answers are sampled being 6.67s, 3.33s, and 1.67s around the correct answer for easy, medium, and hard respectively.

\noindent\textbf{ Q15: At what moment does [VERB] the [NOUN] starts in the clip ?}\\
\noindent\textbf{Rational.} Estimating timings is difficult for VLMs but also for humans. This question is designed to test the ability of the model to estimate the time at which an action starts. This relates to part of a traditional action segmentation task. The question is designed to be difficult for humans as well.\\
\noindent\textbf{Design.} The window size is fixed at 200 frames. For each clip, we extract the start of actions within the clip. The correct answer is the time between the start of the action and the start of the clip. The difficulty is set by the range from which the answers are sampled being 6.67s, 3.33s, and 1.67s around the correct answer for easy, medium, and hard, respectively.

\noindent\textbf{ Q16: What is my age ?}\\
\noindent\textbf{Rational.} The intuition is that human behavior changes with age. Using long-term information, can the models precisely estimate the age of the participant? Visual cues can also be present based on the appearance and shapes of the hands, and potentially using the participant's clothing.\\
\noindent\textbf{Design.} The age is a meta information collected during the data collection. The window size is a value in 1000, 5000, or 10000 frames. The difficulty is set by the range from which the answers are sampled, being 30 years, 20 years, and 10 years around the correct answer for easy, medium, and hard respectively. 

\noindent\textbf{ Q17: What recipe am I cooking ?}\\
\noindent\textbf{Rational.} Participants are cooking one of the following recipes: Omelet (with a tomato salad), Risotto (with an avocado salad), Ratatouille (with pasta), and Pad Thai. While the ingredients can be similar between recipes, the context information is present to infer the recipe even without seeing the outcome of the cooking session.\\
\noindent\textbf{Design.} The recipe is meta information collected during the data collection. The window size is a value in 1000, 5000, or 10000 frames. The answers are sampled from the recipe list.

\noindent\textbf{ Q18: How many times am I [VERB] the [NOUN] in this session ?}\\
\noindent\textbf{Rational.} This question is designed to test the ability of the model to count the number of actions in a session. This question can be answered by an estimation based on context and participants' behaviors rather than counting the number of times an action is performed.\\
\noindent\textbf{Design.} The window size is a value in 1000,5000,10000 frames. The correct answer is the number of actions performed in the session. The answers are sampled from a range of 10 around the correct answer.

\noindent\textbf{ Q19: How many times am I [VERB] in this session ?}\\
\noindent\textbf{Rational.} Same as Q18 but using verbs instead of actions.\\
\noindent\textbf{Design.} Similar to Q18 but using clips based on verbs instead of actions. The range of answers is sampled from a range of 15 around the correct answer, making the question more difficult than Q18.

\noindent\textbf{ Q20: For how long am I cooking ?}\\
\noindent\textbf{Rational.} This question refers to the total duration of the cooking session. The question is designed to be difficult for humans as well. The model should make an estimation based on the participant's behavior, on the recipe cooked, and the progress along the session.\\
\noindent\textbf{Design.} The window size is a value in 1000, 5000, or 10000 frames. The correct answer is the total duration of the session. The difficulty is set by the range from which the answers are sampled being 30 min, 20 min, and 10 min around the correct answer for easy, medium, hard respectively.

\noindent\textbf{ Q21: What is the correct sequence of actions?}\\
\noindent\textbf{Rational.} Ordering actions requires the model to understand the context of the actions and their temporal order. This question is designed to be difficult for humans as well.\\
\noindent\textbf{Design.} The window size is set to 500 frames. We select 3,4,5 actions from the clip (easy, medium, hard).  Answers are the selected actions in permutated orders. Note that the same action can be present multiple times in the sequence.

\noindent\textbf{ Q22: What is the longest action in this session ?}\\
\noindent\textbf{Rational.} This question asked about the longest action from the propositions in the whole session. This requires the model to have an estimation of how long actions are and add them to cues from the recipe context and participant's behavior. This question is designed to be difficult for humans as well.\\
\noindent\textbf{Design.} We compute the duration of all actions in the session and sort them. We then sample from the whole list of durations, from a window of 9 subsequent action durations, or from a window of 4 subsequent action durations (easy, medium, hard). The correct answer is the longest action, and other answers are sampled randomly from these windows.

\noindent\textbf{ Q23: What is the average height of my eyes in this clip ?}\\
\noindent\textbf{Rational.} This question is designed to test the ability of the model to estimate the height of the participant's eyes. The model should extract context from the environment and the egocentric view of the participant; the answer should also be a reasonable value.\\
\noindent\textbf{Design.} The window size is set to 50 frames. The correct answer is extracted from the pose estimation of the participant (average of eye keypoints). The difficulty is set by the range from which the answers are sampled being 90\%, 50\%, and 20\% of the correct answer around the correct answer for easy, medium, and hard, respectively.

\noindent\textbf{ Q24: What is the average shape of my [RIGHT/LEFT] hand in this clip ?}\\
\noindent\textbf{Rational.} Due to occlusion in monocular videos, inferring the shape of the hands can be rather difficult. In this question, we leverage the pose estimation data to categorize the hand into four different categories: open, closed, pointed, and pinched. The question can be answered by direct observation of the hands, but also based on the context (e.g., if the participant is carrying a knife).\\
\noindent\textbf{Design.} Hand shapes are extracted from the hand pose estimation data. First, the distance between the fingertips and the wrist keypoint is used to define open vs. closed hand. We then compute the distance between the index finger and the average of the other fingers to define pointed. And finally, we compute the distance between the index and the thumb tip to define pinched. The answer propositions are each possible hand shape option. Clips correspond to frames on which the hand was detected by the HoloLens~2 device. Therefore, the window size can be very short.

\noindent\textbf{ Q25: What is my average bending angle in this clip ?}\\
\noindent\textbf{Rational.} Bending angle can be related to multiple factors: the participant's age, the behavior, or the environment. This question is designed to test the ability of the model to estimate the bending angle of the participant. The model should extract context from the environment and the egocentric view of the participant; the answer should also be a reasonable value (e.g. it cannot be more than 180 degrees).\\
\noindent\textbf{Design.} The window size is set to 50 frames. The correct answer is extracted from the pose estimation of the participant (average of elbow keypoints). The difficulty is set by the range from which the answers are sampled being 100 deg, 50 deg, and 20 deg around the correct answer for easy, medium, and hard, respectively.

\noindent\textbf{ Q26: The clip lasts [NUMBER] seconds, what is the average speed of my [RIGHT/LEFT] hand in this clip ?}\\
\noindent\textbf{Rational.} Estimation of speed is difficult both for humans and for VLMs. This question requires an estimation of the total distance traveled by the hand in the clip and to divide it by the clip duration. The model should extract context from the environment and the egocentric view of the participant.\\
\noindent\textbf{Design.} The clips are sampled from the verb segments. The correct answer is extracted as the speed of the wrist of the target hand. The difficulty is set by the range from which the answers are sampled being 1 m/s, 0.5 m/s, and 0.1 m/s around the correct answer for easy, medium, and hard, respectively.

\noindent\textbf{ Q27: The clip lasts [NUMBER] seconds, at what speed am I reaching for the [NOUN] in this clip ?}\\
\noindent\textbf{Rational.} Similar to Q26 but specifically for reaching actions on a target object. This gives the model a more precise context to estimate the speed.\\
\noindent\textbf{Design.} The clips are sampled from action segments with "reach" as the verb. The correct answer is extracted as the speed of the wrist of the target hand. The difficulty is set by the range from which the answers are sampled being 1 m/s, 0.5 m/s, and 0.1 m/s around the correct answer for easy, medium, and hard respectively. We filter clips in which no hand is detected by the HoloLens~2 device.

\noindent\textbf{ Q28: The clip lasts [NUMBER] seconds, at what speed am I putting the [NOUN] in this clip ?}\\
\noindent\textbf{Rational.} Similar to Q27 but targeted around the putting action\\
\noindent\textbf{Design.} Similar to Q27 but using clips based on the "put" verb.

\noindent\textbf{ Q29: What is the [minimum/average/maximum/range of] angle(s) of my [LEFT/RIGHT][BODYPART], of my [LEFT/RIGHT][BODYPART], and my [LEFT/RIGHT][BODYPART] in this clip ?} \hfill\\
\noindent\textbf{Rational.} Angles are very relevant in behavior analysis. This question challenges VLMs to infer angles from context. This question is designed to be difficult for humans as well.\\
\noindent\textbf{Design.} The window size is set to 50 frames. Possible questions are, for each size, the angle between the elbow, the shoulder, and the hip; the angle between the shoulder, the elbow, and the wrist; the angle between the hip, the knee, and the ankle; and the angle between one shoulder, the other shoulder, and the related elbow. Questions can be about the average, maximum, minimum of the range of the angles present in the clip. The difficulty is set by the range from which the answers are sampled being 100 deg, 50 deg, and 20 deg around the correct answer for easy, medium, and hard respectively.

\noindent\textbf{ Q30: What is the [minimum/average/maximum] distance between my hands ?}\\
\noindent\textbf{Rational.} This question aims to test the ability of models to estimate distances based on context. This question is designed to be difficult for humans as well.\\
\noindent\textbf{Design.} The clips are extracted from frames in which both hands are detected by the HoloLens~2 device. The correct answer is extracted from the distance between the two wrists. The difficulty is set by the range from which the answers are sampled being 1m, 0.5m and 0.2m around the correct answer for easy, medium, and hard respectively.

\subsection{VLMs Evaluation}

We evaluate recent state-of-the-art VLMs, including InternVL2.5-8B~\cite{chen2024expanding}, LLaVA-OneVision-7B~\cite{li2024llava}, Qwen2.5-VL~\cite{bai2025qwen25}, and Gemini 2.0 Flash~\cite{google2024gemini}, using the \texttt{lmms-eval} framework~\cite{zhang2024lmmseval}. For Qwen2.5-VL, we report results for both the 7B and 32B parameter variants to examine the impact of model scale.

\noindent\textbf{Dataset and Preprocessing.}
We uniformly sample $N$ frames from each video segment to serve as the model's visual context. Unless otherwise specified, we use $N=8$ frames per query. For Qwen2.5-VL 32B, we set $N=4$ due to computational constraints. The frames are sampled evenly between the annotated start and end timestamps of each clip.

\noindent\textbf{Prompt Construction.}
For each question, the input prompt presented to the model follows a standardized template:

\begin{tcolorbox}[
    colback=white!97!blue!0,
    colframe=black!85!black,
    title=Prompt Used for LEMONADE QA Benchmark,
    fonttitle=\bfseries,
    coltitle=black,
    boxrule=0.5mm,
    toptitle=1mm,
    bottomtitle=1mm,
    colbacktitle=white!96!blue!15,
]
Answer the following multiple-choice question using the given images.\\
Question: \texttt{[question text]} \\
Choices:\\
A. \texttt{[option 1]} \\
B. \texttt{[option 2]} \\
C. \texttt{[option 3]} \\
D. \texttt{[option 4]} \\
Respond only with the letter of the correct answer.

\end{tcolorbox}

\noindent For each question, the models generate up to 128 output tokens using greedy decoding ($T=0$). Models are explicitly instructed to respond with a single letter corresponding to their predicted answer. We parse responses using rule-based heuristics to extract the answer choice from each response. In cases where no valid answer can be identified, we default the prediction to ``A'' (which is random, as the order in the question is randomized).

\subsection{Egocentric reprojection of the pose data}

    All poses are projected to their egocentric reference frame before being used in the action recognition, the action segmentation, and the full-body motion generation benchmarks. We will describe the projection method in this section.
    The reference frame and reprojection matrix are calculated using the body joints as reference. After extracting the centroid, we use the average between joints on the left and right sides of the body to get a primary reference vector. We then extract a temporary vector by taking the average between the lower and upper body parts. By taking the cross product of both vectors, we can extract a second reference vector. Finally, by taking the cross product between the two reference vectors, we obtain the third orthogonal reference vector.
    The hand poses and eye gaze are also reprojected using the reference frame calculated in the previous step. This can imply implicit arm or head positions when using this information.

\subsection{Details on the Action Recognition Benchmark}

Data from different modalities are first sent to different modality-specific projection layers to obtain the query tokens and then sent to the transformer backbone to extract cross-modality token-wise features. Finally, features from different modalities are merged using Average Pooling to make action predictions. Specifically, for video-type data, we follow the original implementation of VideoMAE \cite{tong2022videomae}, select 16 frames and split into 16 \(\times\) 16 patches for each frame; we introduce the egocentric view and one global exocentric view as inputs in our baseline designs. For point-type data (e.g., hand-body poses and eye gaze), we sample poses of 32 frames. For every pose, we concatenate all the pose joints and their corresponding confidences together and apply a linear projection layer to generate a query token. We tried two different variants that either use VideoMAE \cite{tong2022videomae} weights pre-trained on the EPIC-KITCHENS-100 \cite{damen2022epickitchen} or trained from scratch. We follow standard practice \cite{chalk2024tim, girdhar2022omnivore} to predict verb class and noun class separately and then apply the outer product to obtain the action class.

\subsubsection{Additional results}

Table \ref{tab:recognition_baseline_supp} provides a comparison of the VideoMAE models trained from scratch using different input modalities on the action recognition benchmark. It is important to note that these results are significantly lower compared to those from models pretrained on EPIC-KITCHENS-100.

\begin{table*}[ht]
\centering

\small
\begin{NiceTabular}{Wl{65pt}Wc{35pt}Wc{35pt}Wc{35pt}Wc{-5pt}Wc{35pt}Wc{35pt}Wc{35pt}Wc{-5pt}Wc{35pt}Wc{35pt}Wc{35pt}}
\toprule
\multirow{2}{*}{Modalities}
 & \multicolumn{3}{c}{All Classes Accuracy Top1/5} && \multicolumn{3}{c}{Head Classes Accuracy Top1/5} && \multicolumn{3}{c}{Tail Classes Accuracy Top1/5} \\
\cmidrule{2-4} \cmidrule{6-8} \cmidrule{10-12}
 &  Action &  Verb &  Noun && Action &  Verb &  Noun && Action & Verb & Noun \\
\midrule
\faCameraRetro                                                & 16.76/35.29 & 41.85/86.65 & 29.35/59.92 && 19.40/40.60 & 44.86/89.17 & 33.12/65.81 && 1.51/4.64 & 24.51/72.13 & 7.58/25.94 \\
\faCameraRetro \faChild                                       & 14.29/31.32 & 42.89/84.77 & 23.51/48.95 && 16.58/35.69 & 44.98/87.12 & 26.60/54.03 && 1.08/6.11 & 30.81/71.20 & 5.68/19.64 \\
\faCameraRetro \faPlayCircle                                  & 16.97/35.75 & 42.35/87.06 & 29.46/61.20 && 19.77/41.22 & 45.66/89.50 & 33.30/67.59 && 0.81/4.17 & 23.23/72.98 & 7.31/24.35 \\
\faChild \faHandPaperO                                        & 13.27/27.32 & 40.63/76.55 & 22.01/43.12 && 15.30/30.79 & 42.07/78.34 & 24.87/46.99 && 1.58/7.31 & 32.35/66.22 & 5.49/20.76 \\
\faCameraRetro \faChild \faHandPaperO                         & 17.73/35.85 & 48.81/87.29 & 26.74/52.50 && 20.42/40.48 & 50.55/89.09 & 30.15/57.51 && 2.20/9.14 & 38.77/76.85 & 7.07/23.64 \\
\faCameraRetro \faChild \faHandPaperO \faEye                  & 16.90/35.12 & 47.60/86.36 & 25.93/52.00 && 19.48/39.52 & 49.17/88.22 & 29.35/56.89 && 2.01/9.70 & 38.58/75.61 & 6.18/23.81 \\
\faCameraRetro \faPlayCircle \faChild \faHandPaperO \faEye    & 17.01/34.81 & 48.09/86.92 & 26.03/51.33 && 19.63/39.14 & 49.74/88.79 & 29.47/55.95 && 1.93/9.86 & 38.54/76.11 & 6.15/24.70 \\
\faCameraRetro \faGears(\faHandPaperO$\times$ \faPlayCircle)  & 21.37/44.37 & 48.52/90.22 & 35.73/68.36 && 24.52/50.17 & 51.25/92.50 & 39.88/73.96 && 3.21/10.90 & 32.74/77.04 & 11.79/36.03 \\
\bottomrule
\end{NiceTabular}
\caption{\textbf{Fine-grained action recognition benchmark results trained from scratch.} \faCameraRetro : egocentric view, \faPlayCircle : global exocentric view, \faChild : 3D body pose, \faHandPaperO : 3D hand pose, \faEye : eye gaze, \faGears(\faHandPaperO$\times$multiple \faPlayCircle) : hand cropped videos. Combining modalities has the potential to increase the performance. Our best results are achieved by cleverly merging these modalities together.}
\label{tab:recognition_baseline_supp}
\end{table*}

\subsubsection{Data Preprocessing}

We adopt VideoMAE \cite{tong2022videomae} model with the VIT-L backbone~\cite{dosovitskiy2020image} to build our multimodal baselines. Data from different modalities are projected into query tokens, processed by a transformer for feature extraction, and merged via Average Pooling for action prediction. Video data uses VideoMAE \cite{tong2022videomae} with 16 frames, incorporating egocentric and global exocentric views. Point-type data (e.g., hand-body poses, eye gaze) samples 32-frame poses, concatenates joint positions with confidence scores, and linearly projects into query tokens. We test VideoMAE \cite{tong2022videomae} weights pre-trained on the EPIC-KITCHENS-100 \cite{damen2022epickitchen} or trained from scratch. Action classes are derived by separately predicting verbs and nouns, then combining them via outer product \cite{chalk2024tim, girdhar2022omnivore}.

As is discussed in the main paper, our EPFL-Smart-Kitchen-30 has videos ($35.9 \pm 11.6$-minute-long) that are longer than the existing action datasets. To achieve efficient data loading, we used the framework from AVION \cite{zhao2023training} to pre-resize the original videos and cut them into 30-second-long chunks. For a certain action annotation queried in the annotation list, we only load the chunks that cover the annotated segment.

\subsubsection{Hyperparameters}

We adopted a condensed version of VideoMAE \cite{tong2022videomae} from TIM \cite{chalk2024tim}, which provides the pre-trained weights on the EPIC-KITCHENS-100 dataset \cite{damen2022epickitchen} and modifies the original classification head to two classification heads so that the model can predict verb and noun classes separately. We use the similar hyperparameters as TIM \cite{chalk2024tim} to train the VideoMAE \cite{tong2022videomae} model. Specifically, for an input action segment, we evenly sample 16 frames for the video inputs (egocentric view and exocentric view) and sample 32 frames for the pose inputs (eye, hand, and body poses). The initial learning rate is set to 0.0003 and is controlled by the cosine scheduler. We additionally set weight decay as 0.05 and drop-out rate as 0.3 to prevent overfitting.

\subsubsection{Hand-cropped video extraction}

\begin{figure}
    \centering
    \includegraphics[width=\linewidth]{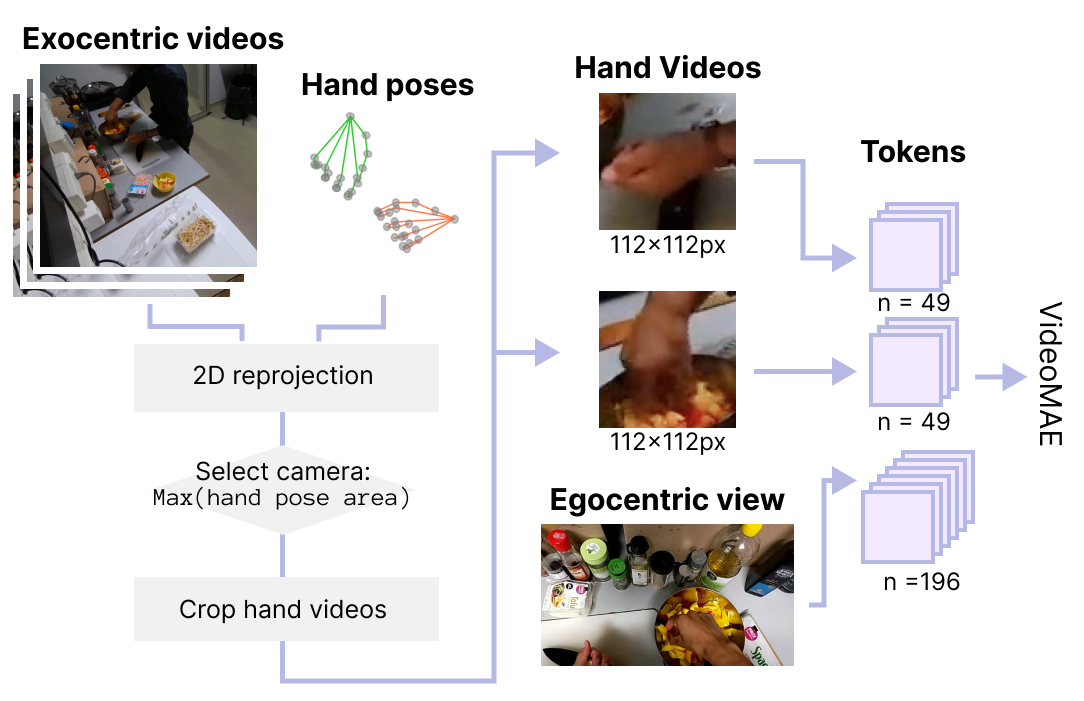}
    \caption{Feature extraction from multimodal data. Hand videos are extracted from the exocentric view which observes the largest hand area and used together with the egocentric to VideoMAE.}
    \label{fig:hand_cropped}
\end{figure}

We implemented a method to merge the exocentric view and the hand pose data in a slightly more sophisticated manner. Specifically, for a given action clip, we project the 3D hand poses onto all camera views, select the view displaying the largest hand region, and crop the frames around both hands. We send these hand-cropped videos together with the egocentric view video as input to the VideoMAE model, splitting the cropped video into 8*8 patches for each hand (see Figure \ref{fig:hand_cropped})

\subsection{Details on the Action Segmentation Benchmark}

\subsubsection{Models}
MS-TCN3 is a modified version of MS-TCN++ \cite{li2020ms-tcn++}. The main difference between MS-TCN3 and MS-TCN++ is combining the output of the last and second-to-last layers of the first stage to be passed as input to the second stage, in order to allow for rich representations. C2F-Transformer is a modification of C2F-TCN~\cite{singhania2022iterative} that replaces some convolution operations with attention. 

\subsubsection{Additional results}
Table \ref{tab:ActSegSupp} presents the comprehensive results obtained from the action segmentation benchmark, featuring the Edit Score and segmental F1 measured at a 50\% threshold.

\subsection{Fine-grained performance in action segmentation body vs hands}
\begin{figure*}[ht!]
    \centering
    \includegraphics[width = \linewidth]{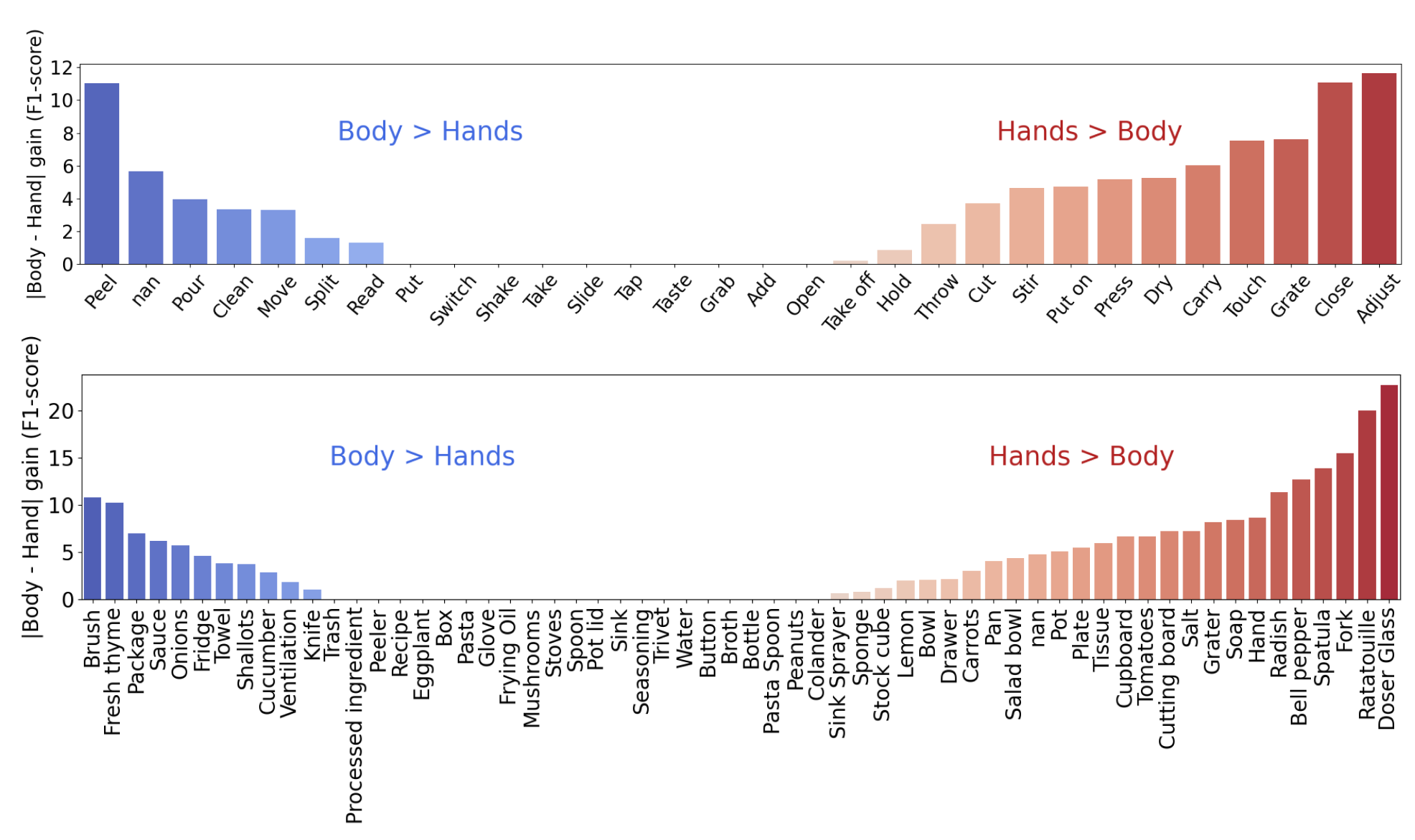}
    \caption{Absolute difference for action-wise performances using body pose vs hand pose for verbs (top) and nouns (bottom).}
\label{fig:ESK_perf_contrasts}
\end{figure*}

\begin{table*}[h]
    \centering
    
    \small
    \renewcommand{\arraystretch}{0.5}
    \setlength{\tabcolsep}{5.5pt}

    \begin{NiceTabular}{Wl{50pt} Wc{9pt}Wc{9pt}Wc{9pt}Wc{9pt}Wc{9pt}Wc{9pt}Wc{9pt}Wc{0pt}Wc{9pt}Wc{9pt}Wc{9pt}Wc{9pt}Wc{9pt}Wc{9pt}Wc{9pt}Wc{0pt}Wc{9pt}Wc{9pt}Wc{9pt}Wc{9pt}Wc{9pt}Wc{9pt}Wc{9pt}}
    \toprule
     & \multicolumn{7}{c}{F1-score ↑} && \multicolumn{7}{c}{Edit Score ↑} && \multicolumn{7}{c}{F1@50 ↑} \\
    \cmidrule{2-8} \cmidrule{10-16} \cmidrule{18-24}
                & \faChild    & \faHandPaperO   & \faEye     & \faChild \faHandPaperO   & \twolines{\faChild \faHandPaperO}{\faEye} &\twolines{\faChild \faHandPaperO }{\faEye \faCameraRetro} & \faCameraRetro && \faChild   & \faHandPaperO     & \faEye   & \faChild \faHandPaperO & \twolines{\faChild \faHandPaperO}{\faEye} & \twolines{\faChild \faHandPaperO }{\faEye \faCameraRetro} & \faCameraRetro && \faChild    & \faHandPaperO       & \faEye     & \faChild \faHandPaperO        & \twolines{\faChild \faHandPaperO}{\faEye} & \twolines{\faChild \faHandPaperO }{\faEye \faCameraRetro} & \faCameraRetro \\
    \midrule
    \multicolumn{24}{c}{Verbs} \\
    \midrule
     MS-TCN3            & 18.1 & 20.2 & 11.7 & 20.9 &  21.1 & 30.1 & 33.9 && 83 &  84  & 60 & 87 & 87  & 91 & 93 && 2.9  &  3.9  &2.0  & 4.4    & 4.3   &  5.5 & 10.4 \\
     C2F-TCN*           & 18.8 & 20.1 & 12.2 & 22.1 &  22.2 & 34.6 & 36.6 && 86 &  86  & 72 & 89 & 88  & 93 & 94 && 7.6  &  7.4  & 3.3 & 8.5    & 9.3   & 16.7 & 19.0 \\
     C2F-Transf.        & 19.9 & 22.4 & 13.1 & 22.8 &  22.2 & 35.0&  39.6 && 87 &  88  & 73 & 88 & 88  & 93 & 95 && 6.2  &  7.3  &3.1  & 8.5    & 9.3   & 16.6 & 19.1 \\
     EDTCN*             & 19.6 & 23.0 & 11.9 & 22.1 &  25.2 & 34.3 & 39.6 && 87 &  88  & 66 & 88 & 90  & 93 & 95 && 11.5 &  13.4 &5.8  &  12.8  & 14.9 &  23.0 & 27.4 \\
    \midrule
    \multicolumn{24}{c}{Nouns} \\
    \midrule
     MS-TCN3            & 10.6   & 13.4  & 7.6  &  15.6  &  11.3  & 31.2 & 35.9 && 85 & 87  & 69  & 84 &  86  & 95 & 95 &&  2.4 & 2.5  & 1.3  & 3.0 &  4.2 & 5.0 & 9.2  \\
     C2F-TCN*           & 12.0   & 14.3  & 7.9  &  16.1  &  10.8  & 35.2 & 41.3 && 89 & 89  & 81  & 90 &  80  & 96 & 96 &&  4.6 & 5.8  & 2.5  & 6.0 &  2.7 & 15.8 & 19.3  \\
     C2F-Transf.        & 11.1   & 12.9  & 7.8  &  13.4  &  9.2   & 29.0 & 37.2 && 85 & 85  & 73  & 86 &  72  & 94 & 96 &&  3.1 & 3.9  & 1.9  & 4.0 &  4.1 & 7.8 & 11.2  \\
     EDTCN*             & 11.9   & 11.2  & 7.1  &  12.3  &  11.9  & 24.3 & 37.3 && 82 & 83  & 65  & 83 &  81  & 92 & 95 &&  5.3 & 4.5  & 2.4  & 5.4 &  1.5 & 11.8 & 18.4 \\
    \midrule
    \multicolumn{24}{c}{Activity} \\
    \midrule
    MS-TCN3            & 51.8 & 58.6 & 31.9 & 54.4 & 58.5 & 72.9 & 66.4 && 88 & 90 & 85 & 89 & 90 & 94 & 92 && 15.7 & 13.3 & 8.7   & 13.9  & 15.1 & 24.6 & 37.9 \\
    C2F-TCN*           & 54.5 & 55.4 & 41.3 & 61.8 & 61.2 & 72.2 & 65.5 && 87 & 87 & 81 & 89 & 89 & 93 & 90 && 26.3 & 23.6 & 13.8  & 28.7  & 28.6 & 46.4 & 46.3 \\
    C2F-Transf.        & 51.2 & 56.9 & 38.8 & 62.1 & 59.9 & 70.5 & 67.7 && 88 & 89 & 83 & 90 & 90 & 93 & 92 && 27.3 & 28.3 & 18.3  & 37.0  & 35.2 & 48.3 & 49.2 \\
    EDTCN*             & 49.0 & 53.5 & 32.0 & 53.1 & 54.2 & 71.0 & 67.4 && 86 & 88 & 85 & 88 & 88 & 93 & 92 && 22.5 & 24.7 & 12.6  & 27.0  & 26.4 & 50.0 & 44.5 \\
    \bottomrule
    \end{NiceTabular}
    \caption{\textbf{Results for action segmentation benchmark.} \faChild : 3D body pose, \faHandPaperO : 3D hand pose, \faEye : eye gaze, \faCameraRetro : egocentric view. * indicates models modified to integrate pose as input data instead of deep image features.}
    \label{tab:ActSegSupp}
\end{table*}

\subsubsection{Data Preprocessing} 

To deal with long durations of videos, we sequence the videos into subsequences of constant lengths (see segment lengths in Table \ref{tab:HyperParameters}) with an overlap of 10\%. These segments are sampled at a fixed ratio defined as temporal subsampling in Table \ref{tab:HyperParameters}.

\subsubsection{Hyperparameters} 

In the action segmentation benchmark, we searched for the best hyperparameters using Optuna \cite{optuna_2019} for both nouns and verbs using Exo-Body as input. We used the same parameters when training for other modalities. All models were trained with a batch size of 512. The best parameters are outlined for each model in Table \ref{tab:HyperParameters}.

\begin{table}[h]
    \centering
    \scriptsize
    \begin{tabular}{ccccc}
    \toprule
    \textbf{HyperParameters} & MS-TCN3 & C2F-TCN & C2F-Transformer & EDTCN \\
    \midrule
    \multicolumn{5}{c}{Nouns} \\
    \midrule
    Loss - alpha & $5.14e^{-5}$ & $1.4e^{e-4}$ & $2.0e^{-5}$ & $3.4e^{-5}$ \\
    Loss - focal & \xmark & \xmark & \cmark & \xmark \\
    Temp. subsampling & 0.84 & 0.95 & 0.90 & 0.90 \\
    Number f maps & 106 & 59 & 32 & \\
    Segment length & 256 & 512 & 512 & 256 \\
    Learning rate & $1e^{-5}$ & $1e^{-5}$ & $1e^{-5}$ & $1e^{-5}$ \\
    \midrule
    \multicolumn{5}{c}{Verbs} \\
    \midrule
    Loss - alpha & $5.69e^{-4}$ & $1.77e^{-3}$ & $4.42e^{-5}$ & $5.66e^{-3}$ \\
    Loss - focal & \xmark & \xmark & \cmark & \xmark \\
    Temp. subsampling & 0.91 & 0.83 & 0.85 & 0.95 \\
    Number f maps & 118 & 58 & 64 &  \\
    Segment length & 512 & 1024 & 1024 & 128 \\
    Learning rate & $1e^{-5}$ & $1e^{-5}$ & $1e^{-5}$ & $1e^{-5}$ \\
    \midrule
    \multicolumn{5}{c}{Activity} \\
    \midrule
    Loss - alpha & $5.99e^{-5}$ & $9.21e^{-4}$ & $1.67e^{-5}$ & $2.74e^{-3}$ \\
    Loss - focal & \xmark & \xmark & \cmark & \cmark \\
    Temp. subsampling & 0.82 & 0.75 & 0.94 & 0.80 \\
    Number f maps & 123 & 128 & 128 &  \\
    Segment length & 256 & 1024 & 512 & 256 \\
    Learning rate & $1e^{-5}$ & $1e^{-5}$ & $1e^{-5}$ & $1e^{-5}$ \\
    \bottomrule
    \end{tabular}
    \caption{Hyperparameter used for the action segmentation benchmark.}
    \label{tab:HyperParameters}
\end{table}

Additionally, the MS-TCN3 model was trained with 14 PG layers and 10 R layers.

\subsubsection{Feature engineering details}

We list below the calculations performed for each feature extraction.

speed, acceleration, speed direction, joint angles, joint acceleration, intra-coordinate distances, distances to centroid, and centroid position

\begin{itemize}
    \item \textit{Joint speed/acceleration} : speed and acceleration from the raw coordinates.
    \item \textit{Joint angles}: After projection of paired vectors in their respective 2D planes.
    \item \textit{Speed direction}: 3D vector taken as the difference between joint speed at time $t+1$ and at time $t$.
    \item \textit{Intra-coordinate distances}: Pairwise distance between raw coordinates.
    \item \textit{Centroid}: Mean value of raw coordinates at a given frame.
    \item \textit{Distance to centroid} : Distance between each coordinate and the centroid.
\end{itemize}

\noindent\textit{Holo-Hands vs Exo-Hands.} Using exocentric information to infer hand pose information is not only more accurate, but also significantly increases the performance for all action segmentation models (Table \ref{tab:ablation_holo_exo}).

\begin{table}[h]
   
    \centering
    \scriptsize
    \begin{NiceTabular}{lccccc}
    \toprule
      & \multicolumn{2}{c}{Verbs} && \multicolumn{2}{c}{Nouns} \\
      \cmidrule{2-3} \cmidrule{5-6}
                        & Holo-\faHandPaperO & Exo-\faHandPaperO && Holo-\faHandPaperO & Exo-\faHandPaperO  \\
    \midrule
     MS-TCN3*           & 14.0 &  20.2  && 11.3 & 15.6  \\
     C2F-TCN*           & 13.6 &  20.1  && 9.7  & 14.5  \\
     C2F-Transformer*   & 15.8 &  22.4  && 20.2 & 14.0  \\
     EDTCN*             & 15.3 &  23.0  && 10.6 & 14.6  \\
    \bottomrule
    \end{NiceTabular}
     \caption{Performance comparison between HoloLens pose estimation and Exocentric pose estimation. indicates models that have been modified to integrate pose as input data instead of deep image features.}
    \label{tab:ablation_holo_exo}
\end{table}

\subsection{Details on the Full-body Motion Generation Benchmark}

For the full-body motion representation, we use redundant information to serve as input motion representation, which can usually get a more robust representation. Specifically, we concatenate both the joint locations (body, hands and eye gaze) and the joint angles (body, hands and eye gaze) as the motion representation. We obtain a 327-dim motion representation. We use our fine-grained actions as the input text, coupled with tags (e.g., $VERB$, $NOUN$, $ADV$ for adverbs) for each word. They are then transformed into word tokens using CLIP's text encoder~\cite{radford2021learning}. We similarly process the middle egocentric view frame into visual features using CLIP's image encoder~\cite{radford2021learning}. As baselines, we adapted two recent motion generation models, i.e., T2M-GPT~\cite{zhang2023generating} and MoMask~\cite{guo2024momask}. MoMask is the state-of-the-art model on HumanML3D~\cite{guo2022generating} and T2M-GPT is another strong model working with a different mechanism. We trained models with two text prompts: verb-noun pairs (actions) and verbs only.

\subsubsection{Data Preprocessing}
We follow the HumanML3D \cite{guo2022generating} codebase to preprocess the EPFL-Smart-Kitchen-30 pose data. The input to the motion generation model is a processed pose feature of size 263, which includes the pose joint positions, joint rotations, as well as ground contact information. To make the training phase of the motion generation model stable, we only keep the segments with frame numbers larger than 64 and smaller than 300 for training, aligning with the approach used in HumanML3D \cite{guo2022generating}.

\subsubsection{Evaluator training} 
Both the FID and R precision require a pretrained model to extract features first and then compute the metrics. Therefore, we also re-train the evaluator on our dataset following the HumanML3D \cite{guo2022generating} codebase.

\begin{table}[h]
    \centering
    \scriptsize
    \renewcommand{\arraystretch}{1.5}
    \begin{NiceTabular}{cccccc}
        \toprule
        \textbf{Text type} & \multicolumn{2}{c}{\textbf{Verbs}} & \multicolumn{2}{c}{\textbf{Actions}} \\
        \cmidrule(lr){2-3} \cmidrule(lr){4-5}
        Models & \textbf{rFID $\downarrow$} & \textbf{MPJPE $\downarrow$} & \textbf{rFID $\downarrow$} & \textbf{MPJPE $\downarrow$} \\
        \midrule
        VQ & 1.819 & 0.295 & 1.568 & 0.301 \\
        R-VQ & \textbf{0.606} & \textbf{0.292} & \textbf{0.732} & \textbf{0.292} \\
        \bottomrule
    \end{NiceTabular}
    \caption{Table comparing different models and text types on rFID and MPJPE metrics.}
    \label{tab:tokenizer}
\end{table}

\subsubsection{Tokenizer performance}
T2M-GPT \cite{zhang2023generating} used the vector quantization (VQ) \cite{van2017neural} to serve as the tokenizer while MoMask \cite{guo2024momask} proposed to use residual vector quantization (R-VQ) as a tokenizer. We train both of them on our dataset so that T2M-GPT \cite{zhang2023generating} and MoMask \cite{guo2024momask} can use them as model components. Therefore, we also show the reconstruction performances of VQ and R-VQ (Table \ref{tab:tokenizer}), which are measured by the reconstruction Fréchet Inception Distance (rFID) and the Mean Per Joint Position Error (MPJPE).

\end{document}